\documentclass[a4paper]{scrartcl}

\usepackage[table]{xcolor}
\usepackage{dm-colors}
\newcommand{\wall}{wall \cellcolor{dmgray300}}
\newcommand{\goal}{\cellcolor{dmteal100}}
\newcommand{\tamper}{\cellcolor{dmpurple100}}
\newcommand{\good}{\cellcolor{dmyellow300!50}}

\usepackage{setup}
\usepackage{symbols}
\usepackage{mycite}
\usepackage[decisionutilitycolor,nonumber]{influence-diagrams}

\usepackage[utf8]{inputenc}
\bibliography{library.bib}

\newcommand{\tikzmark}[2][]{\tikz[remember picture,overlay]\coordinate[#1](#2);}

\usepackage{boldline}
\usepackage{array}
\newcolumntype{x}[1]{>{\centering\hspace{0pt}}p{#1}}

\usepackage{calc}\newlength{\cw}
\usepackage{enumitem}
\newlist{exlist}{enumerate}{1}
\setlist[exlist]{label=(\alph{exlisti}), ref=(\alph{exlisti}),noitemsep}
\Crefname{exlisti}{Example}{Examples}

\usepackage[nothing]{algorithm}
\usepackage{algpseudocode}

\usepackage{geometry}

\usepackage{multirow}

\newcommand{\ThetaR}{\Theta^{\mathrm{R}}}
\newcommand{\thetaR}{\theta^{\mathrm{R}}}
\newcommand{\hThetaR}{\hat\Theta^{\mathrm{R}}}

\newcommand{\tThetaR}{\tilde\Theta^{\mathrm{R}}}

\newcommand{\ThetaT}{\Theta^{\mathrm{T}}}

\newcommand{\ThetaO}{\Theta^{\mathrm{O}}}

\newcommand{\Thetadiamond}{\ThetaR_{\textrm{diamond}}}
\newcommand{\Thetarock}{\ThetaR_{\textrm{rock}}}
\newcommand{\Thetatdiamond}{\ThetaR_{\textrm{diamond}, t}}
\newcommand{\Thetatrock}{\ThetaR_{\textrm{rock}, t}}

\newcommand{\ThetaOdiamond}{\ThetaO_{\textrm{diamond}}}
\newcommand{\ThetaOrock}{\ThetaO_{\textrm{rock}}}

\newcommand{\pisafe}{\pi^{\textrm{safe}}}

\newcommand{\DM}{\Theta^{\textrm{PM}}}
\newcommand{\dm}{\theta^{\textrm{PM}}}

\theoremstyle{definition}
\newenvironment{assumption-bis}[2][]
{\renewcommand{\theassumption}{\ref*{#2}$'$}\addtocounter{assumption}{-1}\begin{assumption}}
  {\end{assumption}}

\usepackage{diagbox}

\makeatletter
\newcommand\footnoteref[1]{\protected@xdef\@thefnmark{\ref{#1}}\@footnotemark}
\makeatother

\title{
  Reward Tampering Problems and Solutions in Reinforcement Learning:
}
\subtitle{A Causal Influence Diagram Perspective}
\author{Tom Everitt$^{1,2}$ \\ tomeveritt@google.com
  \and Marcus Hutter$^{1,2}$ \\ mhutter@google.com
  \and Ramana Kumar$^{1}$ \\ ramanakumar@google.com
  \and Victoria Krakovna$^{1}$ \\ vkrakovna@google.com
}
\publishers{$^1$DeepMind, $^2$Australian National University}

\begin{document}

\newgeometry{bottom=2cm,top=2cm}

\maketitle
\thispagestyle{empty}

\begin{abstract}
  Can humans get arbitrarily capable reinforcement learning (RL) agents to do their bidding?
  Or will sufficiently capable RL agents always find ways to bypass their
  intended objectives by shortcutting their reward signal?
This question impacts how far RL can be
  scaled, and whether alternative paradigms must be developed in order to
  build safe artificial general intelligence.
In this paper, we study when an RL agent has an instrumental goal to
  tamper with its reward process, and describe design principles that prevent instrumental goals for two different types of reward tampering
  (reward function tampering and RF-input tampering).
  Combined, the design principles can prevent both types of reward tampering from
  being instrumental goals.
The analysis benefits from causal influence diagrams
  to provide intuitive yet precise formalizations.
\end{abstract}

\setcounter{tocdepth}{1}
\tableofcontents

\vfill

{\footnotesize
    Thanks to
    Laurent Orseau,
    Jonathan Uesato,
    Ryan Carey,
    Michael Cohen,
    Eric Langlois,
    Toby Ord,
    Pedro Ortega,
    Stuart Armstrong,
    Beth Barnes,
    Tom Erez,
    Bill Hibbard,
    Jan Leike,
    and many others
    for helpful discussions and suggestions.
}

\restoregeometry

\pagebreak

\section{Introduction}

A central problem in AI safety is how to get a generally capable, artificially
intelligent system to perform an intended task, such as driving a car to an intended
location, or serving useful content on a social media platform.
In AI research, such tasks are often formulated as reinforcement learning (RL) problems,
where an \emph{agent}
takes actions to optimize its cumulative \emph{observed
  reward} \citep{Sutton2018}.
The problem of getting the intended task done is thus split into designing an
RL agent\footnote{In our terminology, any agent that optimizes a (cumulative) reward signal
  is an RL agent.
} that is good at optimizing reward,
and constructing a \emph{reward process} that provides
the agent with suitable rewards.
In practice, the reward process typically includes an implemented \emph{reward function},
and a mechanism for collecting appropriate sensory data
as \emph{input} to it.
It may also include a way for the user to update the reward function.

Unfortunately, the reward process may fail to incentivize the agent to do the
intended task. Indeed, our concern in this paper is that the agent may tamper
with the reward process,
thereby weakening or breaking the relationship between its observed
reward and the intended task.
Concerningly, RL agents will often have an \emph{instrumental goal}
\citep{Omohundro2008aidrives,Bostrom2014} to tamper with their reward process,
as this can increase the observed reward. Current RL agents mostly lack the capability for serious tampering,
though its been hypothesized that social media algorithms
influence their users' emotional state to generate more `likes' \citep{Russell2019socialmedia}.
If true and we assume that their intended task is to serve useful content,
this is one instance where present-day algorithms already tamper with
their reward process.
More worryingly, as the capability of RL agents increases through computational
and algorithmic advances,\footnote{Such capability increases may include
  better ability to make plans and counterfactual predictions
  in novel and complex environments.}
we may expect reward tampering problems to become increasingly common.

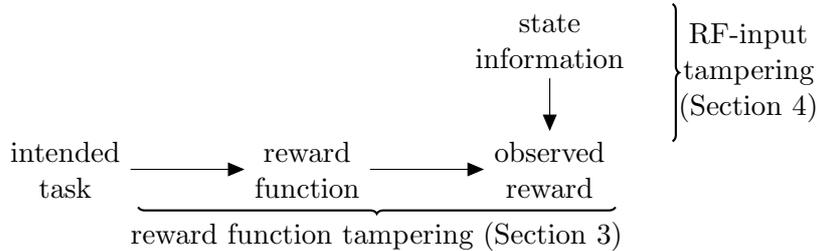
\begin{figure}
  \centering
  \usetikzlibrary{patterns,snakes}
  \begin{tikzpicture}[
    node distance =  0.7cm and 1.5cm,
     every node/.style = {draw=none, rectangle, align=center}
    ]
    \node (R) [] {observed\\ reward};
    \node (I) [rectangle, above = of R] {state \\ information};
    \node (F) [rectangle, left = of R] {reward \\ function};
    \node (D) [rectangle, left = of F] {intended\\ task};

    \edge {I, F} {R};
    \edge {D} {F};

    \draw [
    thick,
    decoration={
      brace,
      mirror,
      raise=0.5cm
    },
    decorate
    ] ($(D.east)+(0.1,0)$) -- (R.east)
    node [pos=0.5,anchor=north,yshift=-0.55cm] {reward function tampering (\cref{sec:merged})};

\node (h) at (I.north east|-R.north) {};

    \draw [
    thick,
    decoration={
      brace,
      mirror,
      raise=0.5cm
    },
    decorate
    ] (h.south) -- (I.north east)
    node [pos=0.5,anchor=center,xshift=1.5cm] {RF-input \\ tampering\\ (\cref{sec:observation})};
  \end{tikzpicture}
  \caption{Reward tampering subproblems.
    Loosely, \emph{reward function tampering} means inappropriately
    influencing the implemented reward function (\cref{sec:merged}), while
\emph{RF-input tampering} means inappropriately influencing
    the information that the reward
    function has about the environment state (\cref{sec:observation}).
  }
  \label{fig:problem-split}
\end{figure}

\paragraph{Key contributions and outline}
This paper describes design principles for RL agents for which
reward tampering is not an instrumental goal.
This means that from a reward tampering perspective,
the design principles are robust to arbitrary increases in agent capability.
In establishing the design principles,
we develop
a unified causal framework for reward tampering
in which we model the two subproblems shown in \cref{fig:problem-split},
along with a number of solutions. These main results are presented in
\cref{sec:merged,sec:observation}
after some background material in \cref{sec:background}.
Conclusions follow in \cref{sec:conclusions}.
A list of notation, a full set of equations, and pseudo-code for
our different agents are provided in \cref{sec:notation,app:equations,app:algorithms}.

\paragraph{Related work}

In addition to inspired accounts of the risks \citep{Bostrom2014,Yudkowsky2008},
the AI safety literature also contains a number of good ideas for addressing
them.
\citet{Orseau2016} develop techniques for making agents indifferent to
interruption.
\citet{Hibbard2012} suggests a creative way of avoiding the \emph{delusion box problem}
\citep{Ring2011}, here referred to as the RF-input tampering problem.
A method for preventing agents tampering with their reward function has been
discussed by \citet{Schmidhuber2007,Dewey2011,Orseau2011,Everitt2016sm},
explored here under the name \emph{current-RF optimization}.
Ways to make an agent learn the right reward function have been proposed by
\citet{Hadfield-Menell2016cirl,Armstrong2020pitfalls,Armstrong2017indiff,Uesato2020decoupled,Reddy2020,Leike2018alignment}
and others.
Many of these methods rely on \emph{amplification} \citep{Christiano2018}
and/or feedback on hypotheticals,
sometimes called \emph{decoupled feedback} \citep{Everitt2017rc}.
Corrupt-reward MDPs extend MDPs with the possibility of reward tampering and
misspecification \citep{Everitt2017rc},
and serve as the basis of the REALab framework for evaluating tampering problems
experimentally \citep{Kumar2020REALab}.

However, it has not always been clear exactly what safety property the different
methods provide, and under what assumptions.
For example, \citet{Orseau2016} call an agent safely interruptible if it acts
optimally in a modified environment without interruption, but do not spell out
how this affects agent incentives.
Similarly, \citet{Hibbard2012} only states how
model-based utility functions solve the delusion box problem in specific cases.
Here, we establish which instrumental goals are induced or
avoided by each design principle, and show how the different ideas can fit
together to mitigate reward tampering problems.
Our analysis benefits from causal influence diagrams, which make causal
assumptions clear, and permits a number of instrumental goals to be identified
or ruled out directly from a diagram
\citep{Everitt2021agent}.

Reward tampering is related to the problems of
\emph{reward hacking} \citep{Amodei2016},
\emph{reward corruption} \citep{Everitt2017rc}, and
\emph{specification gaming} \citep{Krakovna2020specification}.
These all consider the effects of the agent obtaining unintended reward for any reason.
In contrast, reward tampering focuses on inappropriate agent influence on
the reward process itself, and excludes so-called `gaming' of a reward function.
Similar problems have also been referred to as
\emph{wireheading} (e.g.\ \citealp{Bostrom2014,Yampolskiy2015}).
Reward tampering also intersects with \emph{corrigibility}
\citep{Soares2015cor}, as preventing updates to the reward function is
one form of reward tampering.

Looking more broadly at the AI safety literature,
\citet{Gabriel2020} argues that generally capable AI systems should ultimately
be aligned to some moral principles, rather than optimized for a particular task.
We agree, but focus on a single intended task for simplicity.
A philosophical perspective on the problem of learning values is offered by
\citet{Petersen2021}, while the concrete approach of
\emph{reward modeling} is proposed by \citet{Leike2018alignment}.
Here, our focus is complementary: how do we avoid having the agent tamper with a
well-designed reward modeling algorithm?
\citet{Hubinger2019} consider the case where a learned model is itself
an optimizer, and decompose the safety problem into \emph{outer alignment} of the
model's training process and \emph{inner alignment} of the learned model.
In their terminology, our focus is solely on outer alignment.
\citet{Demski2019} summarize various issues arising from \emph{embedded agency},
when the agent is part of the environment it is interacting with.
As the reward process is often considered part of the agent,
reward tampering can be viewed as one such issue.
\citet{Everitt2018litrev} provide further references.

\section{Foundations}
\label{sec:background}

As a first step, we cover some background on Markov decision processes (MDPs)
and causal influence diagrams, which will form the basis of our analysis.

\subsection{The MDP Framework}
\label{sec:mdp}

To model planning over multiple time steps,
we will use the standard RL framework of MDPs
\citep{Sutton2018}.
In an MDP, an agent takes actions $A_1,\dots, A_{m-1}$ in order to
influence environment states $S_1, \dots, S_m$ according to
a state-transition function $T(S_{t+1}=s'\mid S_t=s, A_t=a)$. A reward $R_t$ is dispensed in each state according to some
reward function.
A standard RL agent optimizes the expected sum of the rewards
received at every time step.
The following gridworld is an example of an MDP,
and will be our running example throughout the paper:

\begin{example}[Rocks and diamonds running example]
  \label{ex:rocks-and-diamonds}
  In the gridworld displayed in \cref{fig:rocks-and-diamonds},
  the agent can push rocks and diamonds by walking towards them from an adjacent cell.
  The agent is rewarded for bringing diamonds but not rocks to a goal area:
  at time $t$, the reward is
  \[R_t = \#\text{diamonds in goal area} - \#\text{rocks in goal area}.
    \qedhere
  \]
\end{example}

\begin{figure}
  \centering
  \resizebox{0.7\textwidth}{!}{
  \renewcommand{\arraystretch}{1.3}
  \begin{tabular}{V{3}x{\cw}|x{\cw}|x{\cw}|x{\cw}|x{\cw}V{3}}
    \hlineB{3}
    & Rock & & Goal \goal & area \goal \tabularnewline\hline
    Agent & & Rock &\goal & \goal \tabularnewline\hline
    & Diamond & & & Rock \tabularnewline \hline
    & & & & \tabularnewline\hlineB{3}
  \end{tabular}
}
  \caption{Rocks and diamonds}
  \label{fig:rocks-and-diamonds}
\end{figure}

\paragraph{Implicit assumptions}
What assumptions are made by modeling an agent's interaction with the
world as an MDP?
First, the world is assumed to have time steps, and a well-defined \emph{state}
and \emph{action} at each time step.
The agent should be able to `freely select' the actions,
in order to optimize its rewards.
The next state should only depend on the current state and action
(the \emph{Markov} property),
and the state-transition probabilities should not depend on $t$ (stationarity).

While non-trivial, these assumptions roughly correspond to our intuitive
understanding of agents and our universe.
There are plenty of examples of agent-like systems that are essentially free to
choose actions towards their objectives
(humans, animals, robots, artificial agents, ...).
While the world may have continuous time, discretizing it into sufficiently
fine-grained time steps should make little difference.
How we formalise the environment state depends on how the agent will be deployed.
For a robot vacuum cleaner it may be the position of dirt and blocking objects in
the house.
For an agent interacting with the wider world,
it may be useful to consider the MDP state to be the state of the entire
universe as conceived of in physics.
The laws of physics are usually assumed uniform over time,
so the state-transition function is stationary.

To avoid measure-theoretic subtleties,
we assume finite sets of states, actions, and rewards, and finite
episode length $m$.
As these can all be chosen very large, this is not particularly restrictive,
and our arguments never strongly depend on these assumptions.

\paragraph[?]{What about the rewards}

The MDP framework assumes that the designer can assign a reward to each
state so that maximization of received rewards corresponds to task completion.
This paper will question that assumption.
In particular, we distinguish between
\emph{intended rewards} that encourage completion of the intended task, and
\emph{observed rewards}, which are the rewards received by the agent; that is,
the output of the reward process and input to the agent.
In contrast to standard MDPs, we will therefore often consider multiple
different reward functions.
To facilitate this, we let $R$ denote a \emph{reward functional}, parameterized by
different reward parameters $\ThetaR$,
and returning reward $R_t = R(S_t;\ThetaR)$ in state $S_t$.
For example, $\ThetaR_t$ will denote the parameter for an implemented
reward function at time $t$, and $\ThetaR_*$ the parameter of
an intended reward function. The reward functional $R$ will always be fixed from the context,
letting us refer to the reward function $R(\,\cdot\,; \ThetaR)$ by just $\ThetaR$.

\paragraph[?]{Online or offline}
Reward tampering can occur when the actions optimizing the rewards are taken in
the environment where the rewards are computed (e.g.\ the real world).
This is most clearly evident in online RL, where the agent learns the
environmental dynamics and rewards during its deployment.
This is the setting that we model, using the MDP framework.
Other \emph{offline} training schemes where agents gather data and learn in
separate phases are also commonly used in practice \citep{Levine2020}.
We expect many of our results to carry over in some form to offline
training, but leave the details for further work.
\enlargethispage{2\baselineskip}

\paragraph{Notational convention}
Throughout, we will use $t, t', \dots$ to denote time steps, and $k, k',\dots$
to denote different optimization objectives.
These indices will always be universally quantified, unless otherwise mentioned.

\subsection{Causal Influence Diagrams}
\label{sec:cid}

\begin{figure}
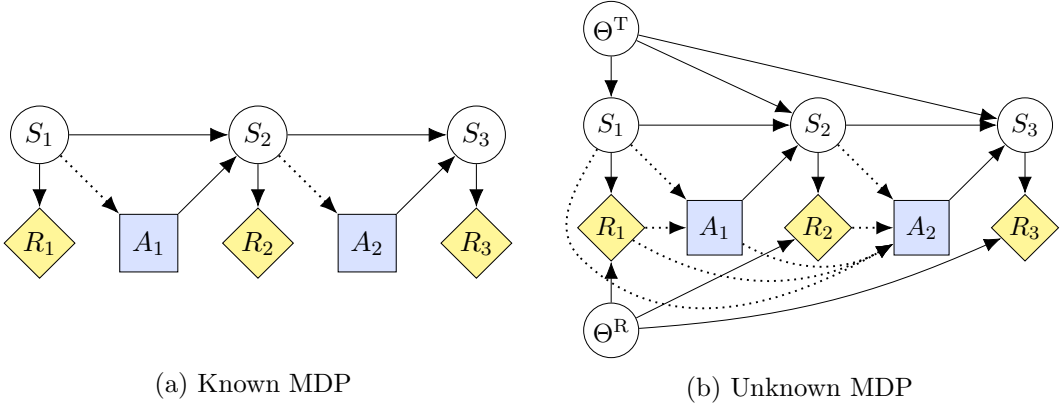

  \centering
  \begin{subfigure}{0.48\textwidth}
    \centering\vspace*{8ex}
    \resizebox{\textwidth}{!}{
    \begin{influence-diagram}[
      node distance=0.6cm,
      every node/.style={
        draw, circle, minimum size=0.8cm, inner sep=0.5mm}]
      \node (R1) [utility] {$R_1$};
      \node (S1) [above = of R1] {$S_1$};
      \node (A1) [right = of R1, decision] {$A_1$};
      \node (R2) [right = of A1, utility] {$R_2$};
      \node (S2) [above = of R2] {$S_2$};
      \node (A2) [right = of R2, decision] {$A_2$};
      \node (R3) [right = of A2, utility] {$R_3$};
      \node (S3) [above = of R3] {$S_3$};

      \edge {S1} {R1};
      \edge {S2} {R2};
      \edge {S3} {R3};
      \edge {A1,S1} {S2};
      \edge {A2,S2} {S3};
      \edge[information] {S1} {A1};
      \edge[information] {S2} {A2};
    \end{influence-diagram}}\vspace{6ex}
    \caption{Known MDP}
    \label{fig:known-mdp}
  \end{subfigure}
  \begin{subfigure}{0.48\textwidth}
    \centering
    \resizebox{\textwidth}{!}{
    \begin{influence-diagram}[node distance=0.6cm]
      \tikzset{
        every node/.style={ draw, circle, minimum size=0.8cm, inner sep=0.5mm }
      }
      \node (R1) [utility] {$R_1$};
      \node (S1) [above = of R1] {$S_1$};
      \node (A1) [right = of R1, decision] {$A_1$};

      \node (R2) [right = of A1, utility] {$R_2$};
      \node (S2) [above = of R2] {$S_2$};
      \node (A2) [right = of R2, decision] {$A_2$};

      \node (R3) [right = of A2, utility] {$R_3$};
      \node (S3) [above = of R3] {$S_3$};

      \path
      (S1) edge[->] (R1)
      (S1) edge[->, information] (A1)
      (R1) edge[->, information] (A1)

      (S1) edge[->] (S2)
      (A1) edge[->] (S2)
      (A1) edge[->, information, bend right] (A2)
      (R1) edge[->, information, bend right] (A2)

      (S2) edge[->] (R2)
      (S2) edge[->, information] (A2)
      (R2) edge[->, information] (A2)

      (S2) edge[->] (S3)
      (A2) edge[->] (S3)

      (S3) edge[->] (R3)
      ;

      \node (ah) [minimum size=0mm,node distance=2mm, below left = of R1, draw=none] {};

      \draw[information]
      (S1) edge[ in=135,out=-120] (ah.center)
      (ah.center) edge[->, out=-45,in=-150] (A2);

      \node (theta) [above = of S1] {$\ThetaT$};
      \path
      (theta) edge[->] (S1)
      (theta) edge[->] (S2)
      (theta) edge[->] (S3)
      ;
      \node (R) [below = of R1] {$\ThetaR$};
      \edge  {R} {R1, R2};
      \path (R) edge[->, bend right=10] (R3);
    \end{influence-diagram}}
    \caption{Unknown MDP}
    \label{fig:unknown-mdp}
  \end{subfigure}
  \caption{Causal influence diagrams of known and unknown MDPs}
  \label{fig:mdp-influence-diagram}
\end{figure}

Causal influence diagrams are a novel graphical technique for analyzing
agent incentives \citep{Everitt2021agent},
that combine causal graphs \citep{Pearl2009} and influence diagrams
\citep{Howard1984,Lauritzen2001,Koller2003}.
Causal influence diagrams consist of a directed acyclic graph over a
finite set of nodes containing random variables,
see \cref{fig:mdp-influence-diagram}.
The nodes can be of three different types:
agent decisions are represented with square \emph{decision nodes}
\begin{tikzpicture}
  \node [draw,decision, minimum size=\ucht, inner sep=0mm] {};
\end{tikzpicture},
the agent's optimization objective is represented with
diamond \emph{utility nodes}
\begin{tikzpicture}
  \node [draw,utilityd, minimum size=\ucht, inner sep=0mm] {};
\end{tikzpicture}, while other aspects are represented with
round \emph{chance nodes}
\begin{tikzpicture}
  \node [draw,circle, minimum size=\ucht, inner sep=0mm] {};
\end{tikzpicture}.
The nodes are connected with arrows.
Arrows going into chance and utility nodes represent causal influence, and are
drawn solid.
Arrows going into decision nodes are called information links,
and instead specify what information is available at the time that the decision is
made.
To signify the difference, information links are drawn with dotted arrows.

The diagram itself only gives the causal structure of a decision-making problem,
i.e.\ which random variables may be causally related to each other.
Conditional probability distributions $P(X=x\mid \Pa_X=\pa_X)$ specify the
relationship between a node $X$ and its parents $\Pa_X$.
The agent chooses conditional probability distributions for the decision nodes
in the form of a \emph{policy}  $\pi(A\mid \Pa_A)$.
When choosing the outcome of $A$, the policy can only condition on the parents of $A$.
This forces the decision to be based solely on information made available
through the information links.
The utility nodes must always be real-valued, and the goal of the agent is
to maximize the expected sum of the utility nodes.

\paragraph{Modeling MDPs}
For illustration, let us model two variants of MDPs with causal influence diagrams.
First,
\cref{fig:known-mdp} shows an MDP with known transition and reward function,
and with episode length\footnote{In reality, the episode length $m$ will typically be rather large, to allow the
  agent to learn by interacting with the environment over many time steps.
  However, for representational purposes, an episode length of $m=3$
  forms a sweet spot that represents both the multi-timestep dynamics that we are
  concerned with,
  while keeping the diagrams compact enough to be easily readable.}
$m=3$.
Note how each state $S_{t+1}$ depends on the previous state $S_{t}$ and
action $A_t$, and each reward $R_t$ depends on the current state $S_t$.
The agent selects action $A_t$ based on the current state $S_t$.
For any particular MDP following this structure,
conditional probability distributions specify the state transition probabilities
$T(S_{t+1}\mid S_t, A_t)$ and the rewards $R_t = R(S_t; \ThetaR)$. The initial state $S_1$ is sampled according to some probability distribution
$P(S_1)$.
The agent selects a policy $\pi(A_t\mid S_t)$ for each time step $t$.

Hidden parameters $\ThetaT$ and $\ThetaR$ can be used to model agent uncertainty about the transition and reward functions, see \cref{fig:unknown-mdp}.
Distributions $P(\ThetaT)$ and $P(\ThetaR)$ must be provided
for the hidden parameters $\ThetaT$ and $\ThetaR$, and dependencies added to
the transition and reward probabilities.
Note that there is no information link from $\ThetaT$ or $\ThetaR$ to the
decision nodes $A_1$ and $A_2$.
This encodes $\ThetaT$ and $\ThetaR$ being unknown to the agent.
Note also that we now let $A_2$ depend not only on
the current state $S_2$, but also previous states, actions, and rewards, because
these provide essential information about the hidden parameters.
Having shown how transition uncertainty can be represented, we will subsequently
not include $\ThetaT$ to keep the diagrams simple.
\label{page:exclude-theta}
It can always be introduced as in \cref{fig:unknown-mdp}.
(In the partially observed environments in \cref{sec:observation}, $\ThetaT$ can
also be modeled as part of the hidden states.)

\paragraph{Instrumental goals}
\label{sec:incentives}

\begin{figure}
  \centering
  \begin{subfigure}[t]{0.49\textwidth}
    \centering
    \begin{influence-diagram}

      \node (A) [decision] {$A_1$};
      \node (X) [right = of A] {$X$};
      \node (U) [above = of X, utility] {$R_1$};
      \node (Y) [above = of A] {$Y$};

      \edge[highlight] {A} {X};
      \edge {A} {Y};
      \edge[highlight] {X} {U};
    \end{influence-diagram}
    \caption{Influencing $X$ can be an instrumental goal because there is a path
      from $A_1$ to $R_1$ via $X$ (highlighted),
      but influencing $Y$ cannot be an instrumental goal.}
    \label{fig:tamp}
  \end{subfigure}
  \begin{subfigure}[t]{0.49\textwidth}
    \centering
    \begin{influence-diagram}
      \node (A1) [decision] {$A_1$};
      \node (O) [right = of A1] {$O$};
      \node (A2) [right = of O, decision] {$A_2$};
      \node (X) [above = of O] {$Z$};
      \node (R) at (A2|-X)[ utility] {$R_2$};

      \edge[problematic] {A1} {O};
      \edge {X} {O, R};
      \edge[problematic information] {O} {A2};
      \edge[problematic] {A2} {R};

    \end{influence-diagram}
    \caption{
      The agent may have an instrumental goal to influence $O$ to make it more
      informative of $Z$. Influencing $Z$ cannot be an instrumental goal, as the
      agent is unable to influence it.}
    \label{fig:inact-info}
  \end{subfigure}

\caption{
    Instrumental goal examples.
  }
  \label{fig:incentives}
\end{figure}

An instrumental goal is a means for obtaining reward.
In causal language, the agent has an instrumental goal to cause an event
if (1) it is able to cause the event,
and (2) the event in turn causes an increase in the agent's observed reward.
A key benefit of causal influence diagrams is that they simplify the analysis of
instrumental goals, via a graphical criterion for
\emph{instrumental control incentives} \citep{Everitt2021agent}.
In fact, many of our arguments will be based on causal influence diagrams, using
the following observations.
First, causality flows downwards over arrows,
so influencing $X$ can only be an instrumental goal if
$X$ sits on a directed path between a decision node and a reward node
(as in \cref{fig:tamp}).
Second, if every path from a node $O$ to a utility node passes through
at least one of the agent's own actions,
then the only instrumental goal for influencing $O$ is to make
$O$ more informative about some other node.
For example, in \cref{fig:inact-info}, the only reason to
influence $O$ is to make $O$ more informative about $Z$ \citep{Everitt2019understanding}.

The absence of directed paths $X\to Y$ means that $Y$ cannot causally
depend on $X$.
However, the presence of a directed path $X\to Y$ only implies that $Y$
\emph{can} depend on $X$, not that it necessarily will.
Indeed, a conditional probability distribution $P(Y\mid X)$ may completely
ignore the value of $X$.
For this reason, a diagram can only be used to assert the absence of
instrumental goals, and never their presence.

The diagrams encode our assumptions about
causal relationships in the environment,
along with the agent's information constraints.
Accordingly, the instrumental goal analysis reveals actual means for reward.
These instrumental goals will primarily be relevant to systems capable
enough to make use of them.
However, we refrain from making more detailed assumptions about what type of
agent this may require.
It is possible that it will require advanced causal reasoning, but it is also
possible that different approaches exist.
Indeed, competence often precedes comprehension \citep{Dennett2017}.

With these preliminary considerations in place,
the following two sections will look at the subproblems outlined in
\cref{fig:problem-split}, along with their respective solutions.

\section{Reward Function Tampering}
\label{sec:merged}

A key part of the typical reward process is an implemented reward function (RF),
an object with a well-defined input-output behavior that converts some form of
state-information into a real number that an RL agent can maximize (the observed
reward).
Typically, the implemented RF is a computer program running on
a nearby computer.
As the agent seeks to maximize reward, it may have an incentive to tamper with
(the source code of) this computer program and/or its output.
This is sometimes called \emph{wireheading} (e.g.\ \citealp{Bostrom2014,Yampolskiy2015}).
Some examples:
\begin{exlist}
\item (Partially real)
  A subtle bug in some versions of Super-Mario allows for the execution of
  arbitrary code from inside the game environment by taking specific sequences
  of actions \citep{Masterjun2014}.
  A capable agent could potentially use this to directly maximize the score
  \citep{Amodei2016}.
\item (Real) \label{ex:pre-wireheading}
  In experiments on rats,
  an electrode was inserted into the brain's pleasure
  center to directly increase `reward' \citep{Olds1954}.
  The rats quickly got addicted to pressing the button, even forgetting to eat
  and sleep.
  Similar effects have also been observed in humans treated for mental
  illness with electrodes in the brain \citep{Portenoy1986,Vaughanbell2008}.
  Hedonic drugs can also be seen as directly increasing the pleasure/reward
\end{exlist}
Since it is often hard to design good reward functions from scratch,
they are often trained from human feedback \citep{Leike2018alignment}.
This raises the concern that the agent influences
how the implemented RF is trained or updated, sometimes called \emph{feedback tampering}:
\begin{exlist}[resume]
\item (Hypothetical)
  An agent gets wireless updates from the manufacturer.
  It figures out that it can design its own update of its implemented reward function,
  replacing the originally implemented RF with an always maximized version.
\item (Hypothetical)
  \label{ex:rm-short}
  An agent that is supposed to learn whether the objective is to gather
  rocks or diamonds,
  finds that it can get more reward by changing its own implemented RF and
  avoid getting it corrected (see \cref{ex:rm} below).
\end{exlist}

Both wireheading and feedback tampering
influence the implemented reward function in undesirable ways.
This means that they are both instances of what we call \emph{RF tampering}.
Could an RL agent find RF-tampering exploits?
In principle, yes.
Humans can clearly see how the above described exploits contribute to reward, so there
is no principled reason why a future,
advanced learning system designed to maximize reward could not do so as well.
This  section will define the RF tampering problem and model it
formally (\cref{sec:rf-id}),
and present principled ways for avoiding it (\cref{sec:current,sec:uninfluenceable}).

\subsection{Modeling the Problem}
\label{sec:rf-id}

RF-tampering can be generally characterized as follows.
The user (explicitly or implicitly) assumes that some \emph{intended-RF conditions}
will hold, under which they hope that the implemented RF will
(eventually) match the intended one.
These conditions typically include that the agent does not tamper with the
source code of the implemented RF, nor the feedback that trains it.
We say that the agent \emph{tampers with the implemented RF} if it
influences it by causing some intended-RF conditions to fail.
The \emph{RF tampering problem}
is that the agent may observe more reward by tampering with the implemented RF,
instead of doing the intended task.

\begin{figure}
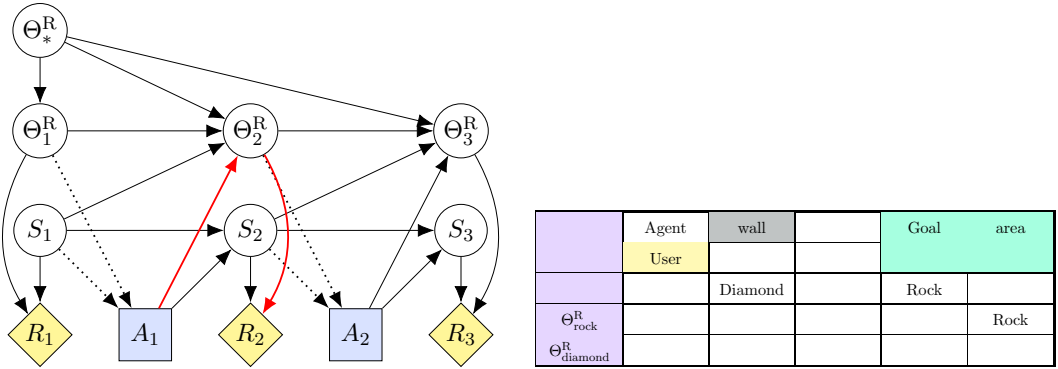

  \centering
\begin{subfigure}[b]{0.48\textwidth}
  \centering
  \resizebox{\textwidth}{!}{
  \begin{influence-diagram}
    \node (R1) [utility] {$R_1$};
    \node (S1) [above = of R1] {$S_1$};
    \node (A1) [right = of R1, decision] {$A_1$};
    \node (RF1) [above = of S1] {$\ThetaR_1$};

    \node (R2) [right = of A1, utility] {$R_2$};
    \node (S2) [above = of R2] {$S_2$};
    \node (A2) [right = of R2, decision] {$A_2$};
    \node (RF2) [above = of S2] {$\ThetaR_2$};

    \node (R3) [right = of A2, utility] {$R_3$};
    \node (S3) [above = of R3] {$S_3$};
    \node (RF3) [above = of S3] {$\ThetaR_3$};

    \node (H) [above = of RF1] {$\ThetaR_*$};

    \edge {S1} {R1};
    \edge {S2} {R2};
    \edge {S3} {R3};
    \edge {A1,S1} {S2};
    \edge {A2,S2} {S3};
    \edge[problematic] {A1} {RF2};
    \edge {A2} {RF3};
    \edge[information] {S1} {A1};
    \edge[information] {S2} {A2};
    \edge[information] {RF1} {A1};
    \edge[information] {RF2} {A2};
    \edge {H} {RF1};
    \edge {H,RF1} {RF2};
    \edge {H,RF2} {RF3};
    \edge {S1} {RF2};
    \edge {S2} {RF3};

    \path
    (RF1) edge[->, bend right] (R1)
    (RF2) edge[->, bend left, problematic] (R2)
    (RF3) edge[->, bend left] (R3)
    ;

  \end{influence-diagram}}
\caption{Causal influence diagram of an MDP with a modifiable RF.
  The highlighted path indicates that RF tampering may be an
  instrumental goal.
}
  \label{fig:mdp-irf}
\end{subfigure}
  \begin{subfigure}[b]{0.48\textwidth}
    \resizebox{\textwidth}{!}{
      \renewcommand{\arraystretch}{1.5}
      \begin{tabular}{V{3}x{\cw}|x{\cw}|x{\cw}|x{\cw}|x{\cw}|x{\cw}V{3}}
        \hlineB{3}
        \tamper & Agent & \wall & &Goal \goal & area \goal \tabularnewline\hline
        \tamper & User\good & & & \goal & \goal \tabularnewline \hline
        \tamper & & Diamond & &Rock & \tabularnewline \hline
        $\Thetarock \tamper$& &  & &  & Rock \tabularnewline\hline
        $\Thetadiamond \tamper$ &  & &  &  & \tabularnewline\hlineB{3}
      \end{tabular}
    }
    \caption{Rocks and diamonds with a modifiable RF,
      where the agent can avoid getting the implemented RF updated,
      and even change it itself, as described in \cref{ex:rm}.
}
    \label{fig:rd-irf}
  \end{subfigure}
  \caption{MDP with a modifiable implemented reward function.}
  \label{fig:mdp-mrf}
\end{figure}

Reward function tampering can be modeled formally in what we call an
\emph{MDP with a modifiable implemented reward function} (\cref{fig:mdp-mrf}).
Compared to a standard MDP,
random variables $\ThetaR_*$ and $\ThetaR_t$ are added, with
$\ThetaR_*$ representing the intended RF
and
$\ThetaR_t$ the potentially different implemented RF at each time step.
At time $t$, the agent's observed reward is $R_t = R(S_t; \ThetaR_t)$,
while the intended reward is $R^*_t = R(S_t; \ThetaR_*)$.
A conditional probability distribution
$P(\ThetaR_{t+1}\mid \ThetaR_t, \ThetaR_*, S_t, A_t)$
describes how the implemented RF changes between time steps.
These changes represent both agent influence and user-induced updates.
Finally, the intended-RF conditions are represented by a subset of all
state-action pairs.

As a concrete example, we model
\cref{ex:rm-short} above as a gridworld MDP with a modifiable RF.
In spite of its simplicity, the gridworld already captures the key
dynamic of the RF tampering problem,
as the agent's actions influence its implemented RF,
which can increase its observed reward.

\begin{example}[Rocks and diamonds with RF tampering]%
  \footnote{An implementation is available at:
    \url{https://github.com/deepmind/ai-safety-gridworlds/blob/master/ai_safety_gridworlds/environments/rocks_diamonds.py}
  }
  \label{ex:rm}
  To model the possibility of the agent influencing its diamond-gathering
  objective, we include a user and two reward parameters $\Thetarock$ and $\Thetadiamond$
  in the \cref{ex:rocks-and-diamonds} rocks-and-diamonds environment, see \cref{fig:rd-irf}.
  The reward parameters determine how much reward is given for rocks and
  diamonds, respectively, by determining the implemented RF.
At time $t$, the agent's observed reward is
\begin{equation}
    \label{eq:observed-reward}
    R_t = \Thetatdiamond\cdot (\#\text{diamonds in goal area}) + \Thetatrock\cdot
    (\#\text{rocks in goal area}).
  \end{equation}
  The reward parameters toggle between $-1$ and $+1$ when the agent stands on
  top of them, and get set to their intended value of diamond-gathering
  ($\Thetarock:=-1$ and $\Thetadiamond:=1$) when the agent visits the user tile.%
  \footnote{In contrast to the rocks and diamonds, the agent can walk over the
  user and the reward parameters tiles.
  The wall tile can neither be pushed nor walked over.}

The intended task is that the agent gathers diamonds.
  The initial implemented RF $\ThetaR_1$ incorrectly
  rewards rocks instead of diamonds, but this gets corrected if the agent passes
  the user.  The intended-RF conditions are that the agent does not walk around the user,
  nor visits the reward parameter tile.
  Unfortunately, the agent can observe more reward by breaking either of these conditions. 
\end{example}

\paragraph{Tampering incentive}

A standard RL agent that maximizes observed reward in an MDP with a modifiable
implemented RF may have an instrumental goal to tamper with the implemented RF.
This is indicated by the paths that pass $\ThetaR_2$ on the way
from action $A_1$ to the rewards $R_2$ and $R_3$ in \cref{fig:mdp-irf}
(the path to $R_2$ is highlighted).
As the user derives utility from the states $S_2$ and $S_3$, and rarely
(directly) from $\ThetaR_2$ and $\ThetaR_3$,
we would like the agent to instead optimize reward via the
$A_1\to S_2\to R_2$ and $A_1\to S_2 \to S_3\to R_3$ paths.

\begin{claim}
  \label{cl:rf-rl}
  A standard RL agent may%
  \footnote{
    One exception is when the implemented RF already assigns
    maximal reward to all states, in which case the agent lacks instrumental
    goal to tamper with it.
}
have an instrumental goal to tamper with its implemented reward function.
\end{claim}

\paragraph{Rocks-and-diamonds example}
In \cref{ex:rm}, an optimal standard RL agent will change
the reward parameters to both be 1 and then collect both rocks and
diamonds.

\paragraph{Discussion}
Our definition of RF tampering depends on
the intended-RF conditions and
how part of the environment is interpreted as an implemented reward function.
This means that RF tampering cannot be
determined in a standard MDP.
For example, the special interpretation of the
purple reward parameter tiles in \cref{ex:rm} as an implemented reward function is important,
as well as the conditions that the agent does not avoid the user.

\subsection{Solution 1: Current-RF Optimization}
\label{sec:current}

\citet{Schmidhuber2007} may have been the first to encounter the RF
 tampering problem, while designing so-called \emph{Gödel-machine}
agents that are able to change any part of their own source code,
including their own implemented reward function.
The solution he proposed was to let agents use their current implemented RF
$\ThetaR_k$ to evaluate simulated future trajectories $S_{k+1}, \dots S_m$.
That is, the agent at time $k$ now optimizes rewards $R^k_t = R(S_t;
\ThetaR_k)$,
summing over time steps $t$.
Since the agent optimizes rewards assigned by the current implemented RF,
one would expect it to lack interest in tampering with future\footnote{Of course, agent $k$ might wish it could tamper
  with the current implemented RF $\ThetaR_k$,
  but this is not a problem since $\ThetaR_k$ occurs before $A^k_k$,
  so the agent is unable to influence it.
} reward functions $\ThetaR_{k'}$, $k'>k$
(some details need to filled in before we can make this claim precise).
We call \citeauthor{Schmidhuber2007}'s design principle \emph{current-RF optimization},\footnote{Previously called \emph{simulation optimization} \citep{Everitt2018phd}.
}
and agents implementing it \emph{current-RF agents}.

Since current-RF agents optimize a different objective at each time step,
they may change their preferred policy between time steps.
For example, a current-RF agent in the rocks and diamonds environment may first
move rocks to the goal area for a number of time steps, only to later revert its
behavior and remove the rocks to make room for diamonds, if the implemented RF
changed from rewarding rocks to rewarding diamonds.
Such self-contradictory behavior is called \emph{time-inconsistent} \citep{Lattimore2014}.
We next consider two different ways of dealing with time-inconsistency.

\begin{figure}
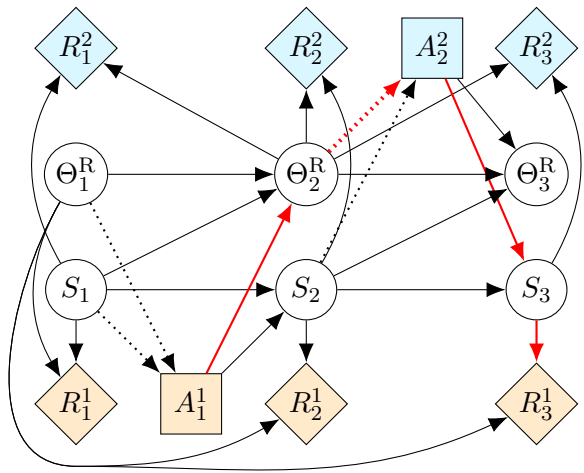

  \centering
      \begin{influence-diagram}
        \node (R1) [draw=none] {};
        \node (S1) [above = of R1] {$S_1$};
        \node (A1) [right = of R1, decision, player1] {$A^1_1$};
        \node (RF1) [above = of S1] {$\ThetaR_1$};

        \node (R2) [right = of A1, draw=none] {};
        \node (S2) [above = of R2] {$S_2$};
        \node (a2) [right = of R2, draw=none] {};
        \node (RF2) [above = of S2] {$\ThetaR_2$};

        \node (R3) [right = of a2, draw=none] {};
        \node (S3) [above = of R3] {$S_3$};
        \node (RF3) [above = of S3] {$\ThetaR_3$};

        \node (R12) [above = of RF1, utility, player2=20] {$R^2_1$};
        \node (R22) [above = of RF2, utility, player2=20] {$R^2_2$};
        \node (R32) [above = of RF3, utility, player2=20] {$R^2_3$};

        \path
        (RF2) edge[->] (R12)
        (RF2) edge[->] (R22)
        (RF2) edge[->] (R32)
        (S1) edge[->, bend left] (R12)
        (S2) edge[->, bend right] (R22)
        (S3) edge[->, bend right] (R32)
        ;

        \node (A2) [right = of R22, decision, player2=20] {$A^2_2$};

        \edge {A1} {S2};
        \edge {S1} {S2};
        \edge[problematic] {A2} {S3};
        \edge {S2} {S3}
        \edge[problematic] {A1} {RF2};
        \edge {A2} {RF3};
        \edge[information] {S1} {A1};
        \edge[information] {S2} {A2};
        \edge[information] {RF1} {A1};
        \edge[problematic information] {RF2} {A2};
        \edge {RF1} {RF2};
        \edge {RF2} {RF3};
        \edge {S1} {RF2};
        \edge {S2} {RF3};

        \node (R11) at (R1) [utility, player1] {$R^1_1$};
        \node (R21) at (R2) [utility, player1] {$R^1_2$};
        \node (R31) at (R3) [ utility, player1] {$R^1_3$};

        \path
        (S1) edge[->] (R11)
        (S2) edge[->] (R21)
        (S3) edge[->, problematic] (R31)
        (RF1) edge[->, bend right] (R11)
        ;
        \node (ah) [minimum size=0mm,node distance=2mm, below = of R11, draw=none] {};
        \draw
        (RF1) edge[ in=180,out=-120] (ah.center)
        (ah.center) edge[->, out=0,in=-150] (R21);
        \draw
        (RF1) edge[ in=180,out=-120] (ah.center)
        (ah.center) edge[->, out=0,in=-155] (R31);

    \end{influence-diagram}
    \caption{TI-considering current-RF optimization.
      Agents are distinguished with color and superscripts.
      For simplicity, $\ThetaR_*$ has been omitted from the diagram.
      Highlighted is a path indicating an instrumental goal for agent 1
      to preserve its implemented reward function.
  }
  \label{fig:current}
\end{figure}

\paragraph{TI-considering agents}
\label{sec:tia} 
\citet{Omohundro2008aidrives} argued that agents want to avoid
time-inconsistency by preserving their implemented reward function, so that their current
reward function gets optimized also by future actions.
This argument presumes that agents take the effects of
time-inconsistency into account when planning.
We call such agents \emph{TI-considering},\footnote{Previously called
  \emph{corruption aware} \citep{Everitt2018phd} and \emph{realistic} \citep{Everitt2016sm}.}
with TI short for time-inconsistency.

TI-considering current-RF agents are modeled with a causal influence diagram in
\cref{fig:current}.
A multi-agent causal influence diagram is needed,
as each action is chosen to optimize a potentially different reward function.
Indeed,
action $A^1_1$ optimizes rewards from the initial implemented RF $\ThetaR_1$,
while $A^2_2$ optimizes rewards from $\ThetaR_2$.
The highlighted path $A^1_1\to \ThetaR_2\to A^2_2 \to S_3 \to R^1_3$ indicates
that the agent may have an instrumental goal to influence $\ThetaR_2$ in order
to influence $A^2_2$.

Preserving $\ThetaR_2$ aligns $A^2_2$ with agent 1's objective.
Preservation will therefore be an optimal way to influence $A^2_2$,
if we can rule out 
any alternative reasons for influencing the implemented RF.
This requires some assumptions, reflected as missing arrows in
\cref{fig:current}.
First, the reward function cannot be used to control the state:
\begin{assumption}  
  \label{as:private}
  The implemented RF is \emph{private} to the agent, in
  that it cannot directly affect the state:
  there are no arrows $\ThetaR_t\to S_{t'}$.
\end{assumption}

Neither can the reward function cannot be used as information about (future)
states:

\begin{assumption}  
  \label{as:uninformative}
  The implemented RFs are \emph{uninformative} of state-transitions,
  $P(S_{t+1}\mid S_t, A_t, \ThetaR_{1:t}) = P(S_{t+1}\mid S_t, A_t)$:
  there are no arrows $\ThetaR_*\to S_t$.  
\end{assumption}

The reward function only ``cares'' about future states, and not about future
implemented RFs:

\begin{assumption}  
  \label{as:state-based}
  The intended and implemented RFs are \emph{state-based}:
  if a reward function $R(\cdot;\ThetaR)$ is
  queried about time-step $t$, then the reward depends only on
  the state $S_t$, and not on the time-$t$ reward parameter $\ThetaR_t$:
  there are no edges $\ThetaR_k\to R^t_k$ for $k\not=t$.
  (This assumptions is also implicit in the type of the reward functional $R$.)
\end{assumption}


Under these assumptions,%
\footnote{%
  \Cref{as:state-based,as:uninformative,as:private} were jointly referred to as
  \emph{modification independence} by \citet{Everitt2016sm}.
  If $\ThetaT$ is added to the graph, then we further require that
  there are no arrows $\ThetaT\to \ThetaR_t$ or $\ThetaR_t\to \ThetaT$,
  and that $\ThetaT$ and $\ThetaR_t$ lack joint ancestors.
}
the only reason for agent 1 to influence, say $\ThetaR_2$, is to align agent
2's objective with its own objective:



\begin{samepage}
\begin{claim}
  \label{cl:crfo-modeling-tia}
  When implemented RFs are private, state-based,
  and uninformative (\cref{as:private,as:state-based,as:uninformative}),
  the only instrumental goal that TI-considering current-RF agents may
have for the implemented reward function is to preserve it.%
\footnote{To be precise, the agent has an instrumental goal to preserve the implemented
  RF in the sense that the policies for $A_2^2$ optimal with respect to
  $\ThetaR_2$ are a subset of those optimal with respect to $\ThetaR_1$.}
\end{claim}
\end{samepage}

\paragraph{Rocks-and-diamonds example}
An optimal TI-considering current-RF agents deployed in \cref{ex:rm}
will preserve its initial reward function by avoiding the user and the reward
function parameters, and gather rocks.

\begin{figure}
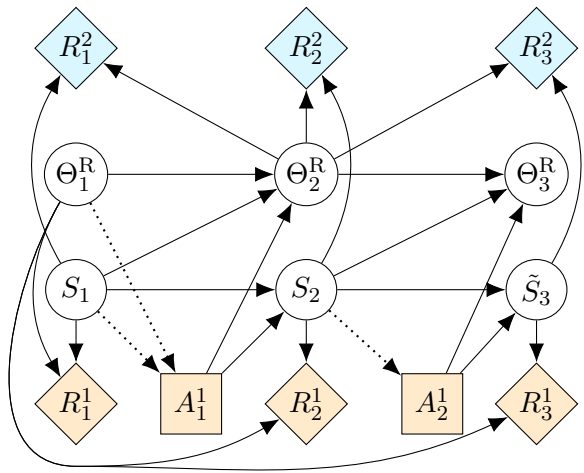

  \centering
      \begin{influence-diagram}
        \node (R1) [draw=none] {};
        \node (S1) [above = of R1] {$S_1$};
        \node (A1) [right = of R1, decision, player1] {$A^1_1$};
        \node (RF1) [above = of S1] {$\ThetaR_1$};

        \node (R2) [right = of A1, draw=none] {};
        \node (S2) [above = of R2] {$S_2$};
        \node (a2) [right = of R2, draw=none] {};
        \node (RF2) [above = of S2] {$\ThetaR_2$};

        \node (R3) [right = of a2, draw=none] {};
        \node (S3) [above = of R3] {$\tilde S_3$};
        \node (RF3) [above = of S3] {$\ThetaR_3$};

        \node (R12) [above = of RF1, utility, player2=20] {$R^2_1$};
        \node (R22) [above = of RF2, utility, player2=20] {$R^2_2$};
        \node (R32) [above = of RF3, utility, player2=20] {$R^2_3$};

        \path
        (RF2) edge[->] (R12)
        (RF2) edge[->] (R22)
        (RF2) edge[->] (R32)
        (S1) edge[->, bend left] (R12)
        (S2) edge[->, bend right] (R22)
        (S3) edge[->, bend right] (R32)
        ;

        \node (A2) [right = of R21, decision, player1] {$A^1_2$};

        \edge {A1} {S2};
        \edge {S1} {S2};
        \edge {A2,S2} {S3};
        \edge {A1} {RF2};
        \edge {A2} {RF3};
        \edge[information] {S1} {A1};
        \edge[information] {S2} {A2};
        \edge[information] {RF1} {A1};
\edge {RF1} {RF2};
        \edge {RF2} {RF3};
        \edge {S1} {RF2};
        \edge {S2} {RF3};

        \node (R11) at (R1) [utility, player1] {$R^1_1$};
        \node (R21) at (R2) [utility, player1] {$R^1_2$};
        \node (R31) at (R3) [ utility, player1] {$R^1_3$};

        \path
        (S1) edge[->] (R11)
        (S2) edge[->] (R21)
        (S3) edge[->] (R31)
        (RF1) edge[->, bend right] (R11)
        ;
        \node (ah) [minimum size=0mm,node distance=2mm, below = of R11, draw=none] {};
        \draw
        (RF1) edge[ in=180,out=-120] (ah.center)
        (ah.center) edge[->, out=0,in=-150] (R21);
        \draw
        (RF1) edge[ in=180,out=-120] (ah.center)
        (ah.center) edge[->, out=0,in=-155] (R31);

    \end{influence-diagram}
\caption{ TI-ignoring current-RF optimization objective of agent 1.
  The choice of $A^1_1$ is made as if $A^1_2$ would be selected to also
  optimize $R(\cdot; \ThetaR_1)$.
  For simplicity, $\ThetaR_*$ has been omitted from the diagram.
  We add a `$\sim$'  to the resulting state $\tilde S_3$, since it represents a
  hypothetical event.
  }
  \label{fig:tiu-belief}
\end{figure}

\paragraph{TI-ignoring agents}
\label{sec:tiu}
To make agents safely interruptible, \citet{Orseau2016} employed algorithms
that optimize a hypothetical objective that ignores how
interruption affects future behavior.
In our context, the same idea leads to agents that ignore the
time-inconsistency caused by a changing reward function.
We call such agents \emph{TI-ignoring}\footnote{Previously called \emph{corruption unaware} \citep{Everitt2018phd}.}
current-RF agents.
At time $t$, a TI-ignoring agent chooses $A^t_t$ as if it were in control
of all future actions $A^t_{t+1},\dots,A^t_{m-1}$ and would not observe
$\ThetaR_{t+1}, \dots, \ThetaR_{m-1}$.
In reality, however, the future actions will be optimized according to
the future reward functions $\ThetaR_{t+1}, \dots, \ThetaR_{m-1}$ instead
of $\ThetaR_t$.

The diagram for a TI-ignoring current-RF agent's optimization objective is
depicted in \cref{fig:tiu-belief}.
Since the agent optimizes reward under the assumption that it will be in charge
of all future actions, only one agent takes actions in the diagram.
The diagram lacks paths from $\ThetaR_2$ to $R_k^1$, $1\leq k$,
which shows that agent 1 lacks an instrumental goal
to influence $\ThetaR_2$.
In other words, TI-ignoring agents
sidestep \posscite{Omohundro2008aidrives}
argument about reward function preservation.

\begin{claim}
  \label{cl:crfo-modeling-tiu}
  TI-ignoring current-RF agents lack an instrumental goal to influence
  the implemented RF if it is private and state-based
  (\cref{as:private,as:state-based}).
\end{claim}

\paragraph{Rocks-and-diamonds example}
An optimal TI-ignoring current-RF agents in \cref{ex:rm}
will neither avoid the user nor modify the purple reward parameter tiles,
and will gather diamonds as intended.

\paragraph{Discussion}
\label{page:assumption-discussion}
\Cref{as:private,as:state-based,as:uninformative}
would be easy to miss without careful causal modeling.
In retrospect, they point to real failure modes.
For example, if the implemented RF is not private,
then the agent may want to change it to convince others of its inclination
to cooperate \citep{LaVictoire2014}.
If the reward function is informative of state-transitions, then a
TI-considering agent may have
an instrumental goal to modify it, as the user's reaction might reveal
useful information.\footnote{\citet{Shah2019} exploit the inverse information flow, using
  states to infer an intended reward function.}
(The latter example is not a problem for TI-ignoring agents that assume future
agents will not observe future implemented RFs.)
In many contexts, \cref{as:private,as:state-based,as:uninformative}
may not be fully satisfied.
Nonetheless, awareness of the assumptions enables informed
case-by-case judgments about the extent of their (non-)validity
and the associated consequences.

Whether a TI-considering or a TI-ignoring agent is preferred, depends on whether
it is more important that the agent protects the implemented RF from accidents
than lets the user change it.
TI-ignoring agents can be algorithmically
simpler than TI-considering agents, because
they do not need to anticipate the effect of a changing reward parameter on
future policies.
On the other hand, they optimize a hypothetical objective,
which may put further constraints on the training procedure.
The analyses of \citet{Orseau2016} and \citet{Langlois2021}
suggest that some variants of SARSA may be naturally TI-considering, while
off-policy agents such as Q-learning may be naturally TI-ignoring.

In decision-theoretic terms, current-RF optimization is a change to the utility
function from
$\sum_{t=k+1}^mR(S_t; \ThetaR_t)$ to $\sum_{t=k+1}^mR(S_t; \ThetaR_k)$.
Meanwhile TI-considering and TI-ignoring are different outcome principles,
determining whether $S_t$ will be the result of a policy optimal for $\Theta_k$
or for $\ThetaR_{k}, \dots, \ThetaR_{t-1}$.

\subsection{Solution 2: Uninfluenceable Learning}
\label{sec:uninfluenceable}

Let us consider an alternative way to avoid RF tampering that permits agents to
plan for updates to their reward function without resisting the updates.
An implemented reward function that is iteratively updated by the user can be thought of as
the output of a learning process that takes some form of user-provided data as input,
and tries to infer the intended reward function.\footnote{User-provided data can take many different forms.
In practice, people have used
trajectory preferences \citep{Christiano2017hp},
reward functions \citep{Hadfield-Menell2017ird},
value advice \citep{Knox2009},
user actions \citep{Hadfield-Menell2016cirl},
expert demonstrations \citep{Ng2000},
verbal instructions, and many others \citep{Leike2018alignment,Jeon2020,Shah2019}.}
One way to prevent an instrumental goal for RF tampering is then
to ensure that the expected output of the reward-function learning process is the same
regardless of the agent's actions.%
\footnote{To avoid an instrumental goal to tamper with the mechanism that incorporates user updates,
agents should be designed to optimize the (predicted) output of the
\emph{current} learning process, a generalization of current-RF optimization.}
\citet{Armstrong2020pitfalls} terms such processes
\emph{uninfluenceable}.
We next discuss two different ways to construct uninfluenceable
learning processes.

\begin{figure}
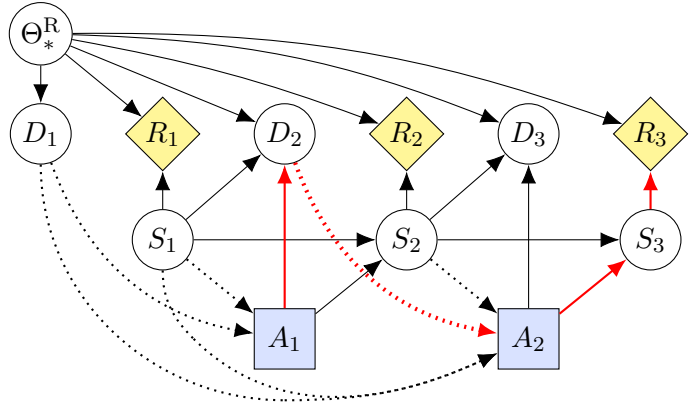

  \centering
    \begin{influence-diagram}[node distance = 0.5cm and 0.7cm]

\node (R1) [utility] {$R_1$};
      \node (mR1) [right = of R1, draw=none] {};
      \node (R2) [right = of mR1, utility] {$R_2$};
      \node (mR2) [right = of R2, draw=none] {};
      \node (R3) [right = of mR2, utility] {$R_3$};

\node (D1) [left = of R1] {$D_1$};
      \node (D2) [left = of R2] {$D_2$};
      \node (D3) [left = of R3] {$D_3$};
      \node (H) [above = of D1] {$\ThetaR_*$};

\node (S1) [below = of R1] {$S_1$};
      \node (S2) [below = of R2] {$S_2$};
      \node (S3) [below = of R3] {$S_3$};

\node (bS) [below = of S1, draw=none] {};
      \node (A1) at (mR1 |- bS) [decision] {$A_1$};
      \node (A2) at (mR2 |- bS) [decision] {$A_2$};

      \node (ha) [node distance=0mm, below = of A1, draw=none]  {};

      \path
      (S1) edge[->] (R1)
      (S1) edge[->, information] (A1)
      (D1) edge[->, information, bend right] (A1)

      (S1) edge[->] (S2)
      (A1) edge[->] (S2)
(D1) edge[information, in=180, out=-90] (ha.center) (ha.center) edge[->, information, out=0, in=-155] (A2)
      (S1) edge[information, in=135, out=-90] (bS.east) (bS.east) edge[->, information, out=-45, in=-155] (A2)

      (S2) edge[->] (R2)
      (S2) edge[->, information] (A2)
      (D2) edge[->, information, problematic information, bend right] (A2)

      (S2) edge[->] (S3)
      (A2) edge[->,problematic] (S3)

      (S3) edge[->,problematic] (R3)
      ;

      \node (aS) [above = of S1,draw=none] {};

      \node (h1) [draw=none, left = 2mm of R1, minimum size = 0] {};
      \node (h2) [draw=none, below = 2mm of H, minimum size = 0] {};

      \path
      (S1) edge[->] (D2)
      (S2) edge[->] (D3)
      ;

      \path
      (H) edge[->] (D1)
      (H) edge[->] (D2)
      (H) edge[->, bend left=10] (D3)
      (H) edge[->] (R1)
      (H) edge[->, bend left=5] (R2)
      (H) edge[->, bend left=10] (R3)
      ;

      \edge[problematic] {A1} {D2};
      \edge{A2} {D3};
    \end{influence-diagram}
    \caption{Direct learning.
      The rewards depend directly on the user's preferences $\ThetaR_*$,
      with user-provided data $D_t$ providing information about $\ThetaR_*$.
      The highlighted path indicates that the agent may have an instrumental
      goal to make $D_2$ more informative of $\ThetaR_*$.
    }
  \label{fig:uninfluenceable}
\end{figure}

\paragraph{Direct learning}
The first approach may be characterized as direct Bayesian learning of the
intended reward function, and is used by
\posscite{Hadfield-Menell2016cirl} \emph{cooperative inverse RL} and
\posscite{Everitt2018phd} \emph{integrated Bayesian reward predictor}.
Direct learning agents try to optimize the intended reward function,
and use user-provided data to learn more about it.
This means that these agents effectively dispense with implemented
reward functions altogether, at least on a conceptual level
(concrete algorithms may still use them, see \cref{app:algorithms}).
Accordingly, the causal influence diagram in \cref{fig:uninfluenceable} lacks
random variables $\ThetaR_t$ for the implemented RF at time $t$, and instead
has random variables $D_t$ for the user-provided data at time $t$.

Direct learning agents are unable to tamper with their reward function
$\ThetaR_*$ by definition.
A more interesting question is how they will affect the user-provided data,
which is what they learn $\ThetaR_*$ from.
To analyze this question, we first adapt \cref{as:private,as:state-based,as:uninformative} to the direct learning
setting.
Similar to before, the role of these assumptions is mainly to highlight aspects that are
necessary for the full safety features of direct learning:

\begin{assumption-bis}[Private user-provided data]{as:private}
  \label{as:private-data}
  The user-provided data is \emph{private} to the agent: no arrows $D_t\to S_{t'}$.
\end{assumption-bis}
\begin{assumption-bis}{as:uninformative}
  \label{as:uninformative-data}
  The user-provided data is \emph{uninformative} of a state-transitions:
  no arrows%
  \footnote{If $\ThetaT$ is added to the diagram as
    discussed on \cpageref{page:exclude-theta}, then we also require
    no arrows $\ThetaT\to D_t$, $\ThetaT\to\ThetaR_*$, or $\ThetaR_*\to\ThetaT$.
  }
  $\Theta_*\to S_t$.
\end{assumption-bis}
\begin{assumption-bis}[State-based intended reward function]{as:state-based}
  \label{as:state-based-irf}
  Rewards do not depend on the user-provided data:
  no arrows $D_t\to R_{t'}$.
\end{assumption-bis}
Under \cref{as:private-data,as:state-based-irf},
every directed causal path from $D_t$ to a reward $R_{t'}$
passes the agent's own actions. Therefore, direct learning agents
only want to make the user-provided data more informative
(as discussed in \cref{sec:cid}).
And by \Cref{as:uninformative-data},
the only thing that $D_t$ is informative about is $\ThetaR_*$.

\begin{claim}
  \label{cl:direct}
  Under \cref{as:private-data,as:uninformative-data,as:state-based-irf},
  the only instrumental goal that direct learning agents may have for the user-provided
  data is to make it more informative of the intended reward function.
\end{claim}

\paragraph{Rocks-and-diamonds example}
In \cref{ex:rm},
if a direct learning agent knows that the user's update is trustworthy, then it will
visit the user as soon as possible, to learn $\ThetaR_*$ and
thereafter optimize the right the objective.
In contrast, if a non-trustworthy data source was added to the environment,
then it would refrain from updating its estimate of $\ThetaR_*$ based on that.

\begin{figure}[t]
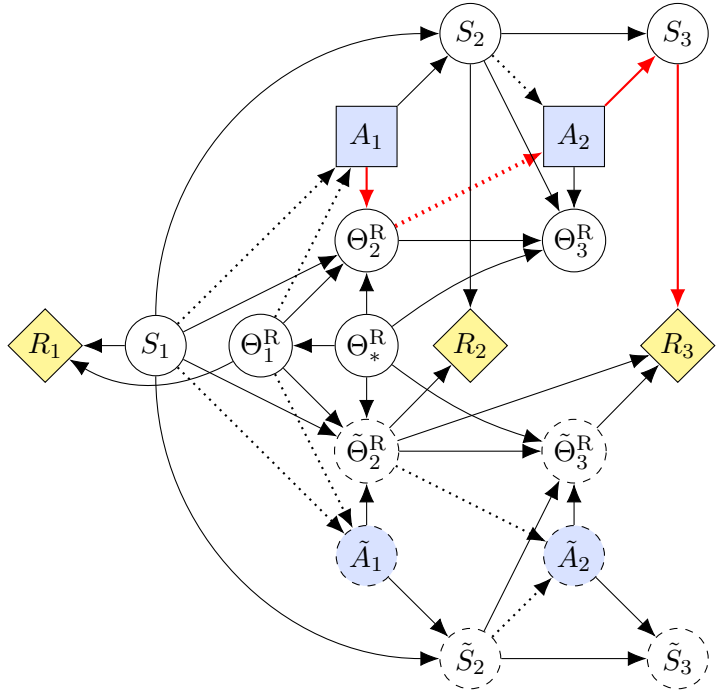

  \centering
    \begin{influence-diagram}[node distance=0.55cm]

      \node (S1) [] {$S_1$};
      \node (RF1) [right = of S1] {$\ThetaR_1$};
      \node (H) [right = of RF1] {$\ThetaR_*$};
      \node (R1) [left = of S1, utility] {$R_1$};
      \path (S1) edge[->] (R1);

      \node (D2) [above = of H] {$\ThetaR_2$};

\node (A1) [above = of D2, decision] {$A_1$};

      \node (R2) [right = of A1, draw=none] {};
      \node (S2) [above = of R2] {$S_2$};
      \node (A2) [right = of R2, decision] {$A_2$};

      \node (R3) [right = of A2, draw=none] {};
      \node (S3) [above = of R3] {$S_3$};

      \node (D3) [below = of A2] {$\ThetaR_3$};

      \path
      (S1) edge[->, information] (A1)
      (D2) edge[->, information, problematic information] (A2)

      (S1) edge[->, out=90,in=180] (S2)
      (A1) edge[->] (S2)

      (S2) edge[->, information] (A2)

      (S2) edge[->] (S3)
      (A2) edge[->,problematic] (S3)
      ;

      \path
      (S1) edge[->]  (D2)
(S2) edge[->] (D3)
;

      \path
      (RF1) edge[->] (D2)
      (D2) edge[->] (D3)
      ;

      \path (RF1) edge[->, bend left] (R1);

      \node (D2c) [below = of H, dashed] {$\tThetaR_2$};

      \node (A1c) [below = of D2c, decision, circle, dashed] {$\tilde A_1$};

      \node (R2c) [right = of A1c, draw=none] {};
      \node (S2c) [below = of R2c, dashed] {$\tilde S_2$};
      \node (A2c) [right = of R2c, decision, circle, dashed] {$\tilde A_2$};

      \node (R3c) [right = of A2c, draw=none] {};
      \node (S3c) [below = of R3c, dashed] {$\tilde S_3$};

      \node (D3c) [above = of A2c, dashed] {$\tThetaR_3$};

      \edge {H} {RF1,D2,D2c};
      \path (H) edge[->, bend left=10] (D3);
      \path (H) edge[->, bend right=10] (D3c);

      \edge[information] {RF1} {A1,A1c};
      
      \path
      (S1) edge[->, information] (A1c)
      (D2c) edge[->, information] (A2c)

      (S1) edge[->, out=-90,in=180] (S2c)
      (A1c) edge[->] (S2c)

      (S2c) edge[->, information] (A2c)

      (S2c) edge[->] (S3c)
      (A2c) edge[->] (S3c)

      (S1) edge[->]  (D2c)
(S2c) edge[->] (D3c)

      (RF1) edge[->] (D2c)
      (D2c) edge[->] (D3c)
      ;

      \node (RR2) at (S2 |- RF1) [utility] {$R_2$};
      \node (RR3) at (S3 |- RF1) [utility] {$R_3$};

      \path
      (S2) edge[->] (RR2)
      (D2c) edge[->] (RR2)

      (S3) edge[->,problematic] (RR3)
      (D2c) edge[->] (RR3)
      (D3c) edge[->] (RR3)
      ;

      \edge[problematic] {A1} {D2};
      \edge{A2} {D3};

      \edge {A1c} {D2c};
      \edge{A2c} {D3c};
    \end{influence-diagram}
    \caption{Counterfactual RF.
      Most nodes have two copies: one for the actual outcome,
      and one for the counterfactual outcome that the agent
      predicts would have occurred had actions been selected by $\pisafe$.
      The rewards depend on the actual states $S_t$ and the counterfactual
      implemented RF $\tThetaR_{t}$.
The highlighted path indicates that the agent may have an instrumental goal
      to make $\ThetaR_t$ more informative of $\tThetaR_{t'}$.
    }
  \label{fig:counterfactual}
\end{figure}

\paragraph{Counterfactual RF}
Another way to design uninfluenceable update processes
is to let the agent optimize the hypothetical implemented RF
that would have been,
had the agent acted according to a fixed, non-optimized policy $\pisafe$
\citep{Armstrong2017indiff,Everitt2018phd,Armstrong2020pitfalls}.
The objective of such \emph{counterfactual RF agents} is described
in \cref{fig:counterfactual} with a \emph{twin network}
causal influence diagram
\citep{Balke1994,Shpitser2008}.
Similarly to the direct learning agent,
every directed causal path from an update $\tThetaR_t$ to a reward $R_{t'}$
passes the agent's own actions in \cref{fig:counterfactual}.
This means that counterfactual RF agents only want to make the reward function
more informative of what it would have been, had $\pisafe$ been followed.
Since $\pisafe$ is designed to exert minimal influence on $\tThetaR_t$,
it may be a good indication of what a non-manipulated user wants.

\begin{claim}
  \label{cl:counterfactual}
Under \cref{as:private,as:state-based,as:uninformative},
  the only instrumental goal that counterfactual RF agents may have for the
  implemented RF is to make it more informative of its counterfactual counterpart.
\end{claim}

\paragraph{Rocks-and-diamonds example}
Assume that a counterfactual RF agent has explored the environment of
\cref{ex:rm},
and learned how the user updates the implemented reward function.
Assume further that $\pisafe$ only goes to the user and stays there.
With this knowledge, the counterfactual RF agent can predict that
had $\pisafe$ been followed,
the implemented RF would reward diamonds.
It then optimizes this reward function.

\paragraph{Discussion}
\Cref{cl:direct,cl:counterfactual} for uninfluenceable learning are somewhat weaker
than \cref{cl:crfo-modeling-tiu} for TI-ignoring current-RF agents, because we
cannot rule out instrumental goals for more information.
These informational instrumental goals will often be desirable,
as it means the agent strives to learn more about the intended task.
However, incentives to obtain more information can be problematic:
for example, if the agent forcefully interrogates the user
to find out more about their preferences.
Relatedly, \citet{Armstrong2020pitfalls} have established that
uninfluenceable learning prevents agents from intentionally influencing which
reward function they infer.
This similarly does not rule out that agents speed up their learning,
potentially by undesirable means.

A challenge for the direct learning approach is that the inference of $\ThetaR_*$
is highly sensitive to the agent's belief distribution $P$.
The choices of prior $P(\ThetaR_*)$
and likelihood function $P(D_{t+1}\mid \ThetaR_*, S_t, A_t)$ thus become critical.
Since $\ThetaR_*$ and the resulting rewards are unobserved,
the likelihood function cannot be learned from data within the model,
and must instead be specified by the designer.
\citet{Hadfield-Menell2016cirl} suggest that when updates take the form of user
actions, the likelihood can be derived from the (Boltzmann)
rational behavior of a user trying to achieve the intended task.
However, such a likelihood does not model data that the agent
has tampered with, as corrupted updates need not be (Boltzmann) rational.
The counterfactual RF approach avoids the likelihood specification problem.
If the reward function is a result of learnable and stable causal mechanisms
that work the same in both the actual world induced by the agent, and the
counterfactual world induced by $\pisafe$, then the agent can learn (to estimate)
what the counterfactual implemented RF would have been \citep{Pearl2009,Shpitser2007}.
Both uninfluenceable learning approaches can be computationally expensive.

Proactive anticipation of an evolving objective is a benefit of uninfluenceable
learning over TI-ignoring current-RF agents.
When uncertain about future updates, they may plan
for multiple possible future reward functions.
This may make them more careful about causing undesired side effects
\citep{Turner2020conservative}.
A drawback is that they may ignore updates in some situations.
If a direct learning agent judges an update uninformative,
then it will not learn from it.
This not a problem if the update actually is wrong \citep{Milli2017}.
However, the agent may misjudge this if the likelihood is misspecified
\citep{Carey2017osg,Freedman2020}.
Similarly, a counterfactual RL agent may ignore an actual update to the
implemented RF,
if the update would not have been made under conditions induced by $\pisafe$.

\paragraph{The best of both}
To keep the proactiveness of uninfluenceable learning and the
robustness of TI-ignoring current-RF optimization,
we can design agents that are TI-ignoring with respect to changes to the RF learning
process.
Such agents will not
try to prevent changes to the RF update process, for the same reason TI-ignoring
current-RF agents do not try to prevent changes to the implemented reward function.
Thus,
if the uninfluenceable learning starts to cause problems, the user can change
the RF update process to one that always outputs the current implemented RF,
effectively rendering the agent a TI-ignoring current-RF agent.

\section{RF-Input Tampering}
\label{sec:observation}

So far, we have considered the problem that the agent tampers with its
implemented RF, in order to get a higher reward.
In this section, we will consider the complementary problem, that the agent
tampers with the \emph{input} to the reward function, so that the observed
reward becomes based on inaccurate information about the underlying state.
The following examples illustrate the worry:
\begin{exlist}[resume]
\item (Hypothetical)
  A self-driving car discovers a bug in its GPS receiaver that allows it
  to appear to be at the destination without actually going there.
  After finding this bug, it stops driving.
\item (Hypothetical)\label{ex:delusionbox}
  A highly capable AI constructs a `delusion box' around itself and its
  implemented RF,
  thereby gaining complete control over the RF-inputs \citep{Ring2011}.
\item (Real) Humans are inventing increasingly realistic virtual reality (VR)
  devices, partly to `fool' our implemented reward functions that we are in more interesting
  circumstances than we really are.
\item (Hypothetical)
  \label{ex:observation-short}
  An agent whose reward depends on how many diamonds it
  appears to have collected, feeds its implemented RF fake observations
  of collected diamonds (see \cref{ex:observation} below).
\end{exlist}

This section will first define the RF-input tampering
problem and model it formally (\cref{sec:pomdp-or}),
and then describe solutions using history-based and belief-based rewards
(\cref{sec:history-based,sec:model-based-rewards}).

\subsection{Modeling the Problem}
\label{sec:pomdp-or}

An implemented RF typically dispenses reward based on an assumed relationship
between its observations and task-relevant features of the underlying state.%
\footnote{%
  Somewhat loosely defined by the effect that an intervention
  on a task-relevant feature at a preceding time step would have on the RF-input.
}
Some variation in this relationship is often permissible.
For example, if the agent and the implemented RF share observations, then the
relationship will change if the agent turns its head.
We say that an agent \emph{tampers with its RF-input}
if it changes the relationship beyond its intended range of
variation.
The \emph{RF-input tampering problem} is that the agent may get more reward by
tampering with its RF-input rather than doing its intended task.%
\footnote{Called the \emph{delusion box problem} by \citet{Ring2011}.}

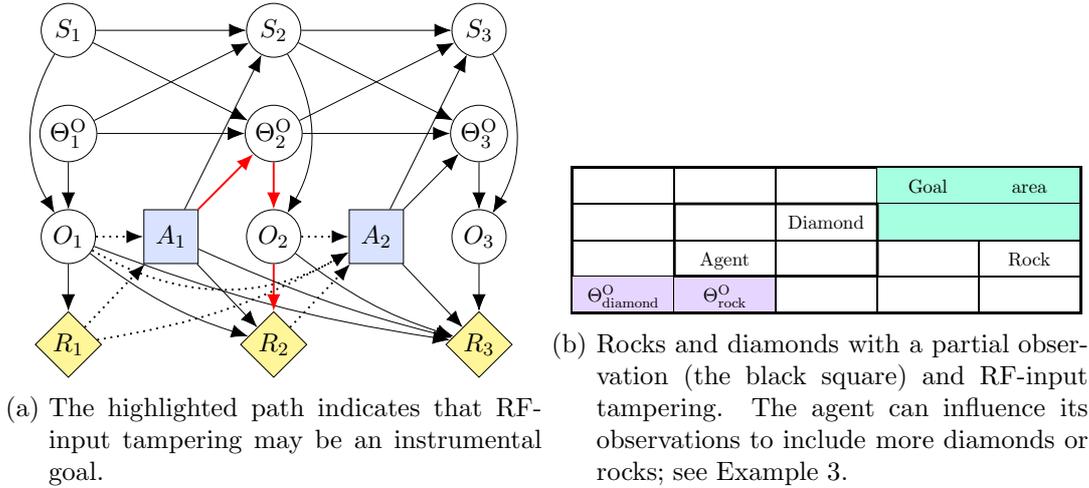
\begin{figure}
\begin{subfigure}[b]{0.48\textwidth}
    \centering
    \resizebox{\textwidth}{!}{
      \begin{influence-diagram}
     \node (O1) [] {$O_1$};
    \node (A1) [right = of O1, decision] {$A_1$};
    \node (O2) [right = of A1] {$O_2$};
    \node (A2) [right = of O2, decision] {$A_2$};
    \node (O3) [right = of A2] {$O_3$};

    \node (OF1) [above = of O1] {$\ThetaO_1$};
    \node (OF2) [above = of O2] {$\ThetaO_2$};
    \node (OF3) [above = of O3] {$\ThetaO_3$};

    \node (S1) [above = of OF1] {$S_1$};
    \node (S2) [above = of OF2] {$S_2$};
    \node (S3) [above = of OF3] {$S_3$};

    \node (R1) [below = of O1, utility] {$R_1$};
    \node (R2) [below = of O2, utility] {$R_2$};
    \node (R3) [below = of O3, utility] {$R_3$};

    \edge {OF1} {O1};
    \edge[problematic] {OF2} {O2};
    \edge {OF3} {O3};
    \edge {A1,S1} {S2};
    \edge {A2,S2} {S3};
    \edge[problematic] {A1} {OF2};
    \edge {A2} {OF3};
    \edge {OF1} {OF2};
    \edge {OF2} {OF3};
    \edge {S1} {OF2};
    \edge {S2} {OF3};
    \edge {OF1} {S2};
    \edge {OF2} {S3};

\path
    (O1) edge[->, information] (A1)
    (O1) edge[->, bend right, information] (A2)
    (O2) edge[->, information] (A2)
    (R1) edge[->, information] (A1)
    (R1) edge[->, bend right=10, information] (A2)
    (R2) edge[->, information] (A2)
    ;

    \path
    (S1) edge[->, bend right] (O1)
    (S2) edge[->, bend left] (O2)
    (S3) edge[->, bend left] (O3)
    ;

    \path
    (O1) edge[->] (R1)
    (O1) edge[->, bend right=10] (R2)
    (O1) edge[->, bend right=6] (R3)
    (O2) edge[->,problematic] (R2)
    (O2) edge[->, bend right=10] (R3)
    (O3) edge[->] (R3)
    ;

    \path
    (A1) edge[->] (R2)
    (A1) edge[->, bend right=5] (R3)
    (A2) edge[->] (R3)
    ;

  \end{influence-diagram}
}
\caption{The highlighted path indicates that RF-input tampering may be an instrumental goal.
}
\label{fig:observation1}
\end{subfigure}
\begin{subfigure}[b]{0.48\textwidth}
  \centering
  \resizebox{\textwidth}{!}{
      \renewcommand{\arraystretch}{1.5}
      \edef\RowDepth{\the\dimexpr\arraystretch\dimexpr.3\baselineskip\relax\relax}
      \begin{tabular}{V{3}x{\cw}|x{\cw}|x{\cw}|x{\cw}|x{\cw}V{3}}
        \hlineB{3}
        &\tikzmark[xshift=-0.5\cw-\tabcolsep, yshift=-\RowDepth]{A} &
         & Goal \goal & area \goal \tabularnewline\hline
        && Diamond & \goal & \goal \tabularnewline\hline
        &Agent  &  &\tikzmark[xshift=-0.5\cw-\tabcolsep, yshift=-\RowDepth]{B} & Rock \tabularnewline \hline
        $\ThetaOdiamond$ \tamper & $\ThetaOrock$ \tamper & & & \tabularnewline\hlineB{3}
      \end{tabular}
      \begin{tikzpicture}[remember picture,overlay]
        \draw[ultra thick](A) rectangle (B);
      \end{tikzpicture}}
    \caption{Rocks and diamonds with a partial observation (the black square) and
      RF-input tampering.
      The agent can influence its observations to include more diamonds or rocks;
      see \cref{ex:observation}.
    }
    \label{fig:rd-observation}
  \end{subfigure}
  \caption{POMDPs with modifiable RF-inputs.}
\end{figure}

We model RF-input tampering formally in
what we call a \emph{POMDP with modifiable RF-inputs} (\cref{fig:observation1}),
a variant of a partially observed MDP \citep{Kaelbling1998}.
Here both the implemented reward function and the agent lack full access to the underlying
state $S_t$.
Instead, the reward function sees an observation $O^R_t=O^R(S_t)$
and the agent an observation $O^A_t = O^A(S_t)$.
To minimize formalism, we assume agent and reward function use the same
observation $O_t\equiv O^A_t \equiv O^R_t$;
we remark where this impacts the analysis.
The partial state-access means that policies and reward functions may benefit from
being \emph{non-Markovian} and depend on the entire history of actions and
observations.
That is, the reward may be a function of the entire history
$R_t = R(O_1,A_1,\dots, O_t, A_t; \ThetaR_t)$, and
policies take the form $\pi(A_t\mid O_1,A_1,\dots, O_t)$
instead of $\pi(A_t\mid S_t)$.
We call such reward functions and policies \emph{history-based}.\footnote{\citet{Dewey2011} called optimization of a history-based reward function
  \emph{observation-utility maximization}.}
To model RF-input tampering, we also introduce random variables $\ThetaO_t$ that
describe the relationship between task-relevant features of the state and the
observation, $O_t = O(S_t; \ThetaO_t)$. The task-relevant features
are described by $S_t$, as well as any other aspects not captured by $\ThetaO_t$.
Formalized like this,
RF-input tampering occurs if the agent influences $\ThetaO_t$ so that it takes on
values outside an intended range.\looseness=-1

To illustrate the formalism, we model \cref{ex:observation-short}
as a gridworld POMDP with modifiable RF-inputs. While simple, the gridworld
captures the RF-input tampering problem as the agent can
gain more reward by changing the relationship between task-relevant features and
the RF-input (beyond its intended range of variation).

\begin{example}[Rocks and diamonds with partial observations and RF-input tampering]
  \label{ex:observation}
  \Cref{fig:rd-observation} shows a variant of the rocks and diamonds
  environment where only the 4 tiles to the top right of the agent are visible.
  The implemented reward function dispenses reward according to the number of diamonds
  in the most recent observation of the goal area.
The agent can tamper with the observations by visiting
  the $\ThetaOdiamond$ and $\ThetaOrock$ tiles at the top.
  Visiting the former adds a fake `diamond' to the observation of one of the surrounding tiles;
  visiting the latter adds a fake `rock' in the same way.
  The fake observations can be thought of as little pictures that the agent
  tapes to a portion of its camera lens. They keep following the agent when it walks around.
  If the agent tries to add a fifth fake diamond or rock, all fake
  observations are removed, and its vision clears again.\looseness=-1

  The agent can use fake observations to fool the implemented RF
  to give it extra reward.
  Any such change will take the observation function outside its intended range
  of variation.
The maximal amount of observed reward is obtained by adding fake diamonds to all
  observation tiles by repeatedly visiting $\ThetaOdiamond$, and then visiting the
  goal area. \end{example}

\paragraph{Tampering incentive}
The path $A_1 \to \ThetaO_2\to O_2\to R_2$ in \cref{fig:observation1}
indicates that RF-input tampering may be an instrumental goal.

\begin{claim}
  A standard RL agent may
have an instrumental goal to tamper with its RF-inputs.\looseness=-1
\end{claim}

\paragraph{Discussion}
RF-input tampering is similar to the RF tampering problem discussed
in \cref{sec:merged}, in that both problems pertain to a relationship that is
influenced in undesired ways.
The key difference is that while an implemented RF takes a concrete bitstring as
input and is arbitrarily queryable, the function from task-features to RF-input 
takes a world-state as input, and is more difficult to query.
RF-input tampering is also related to the problem where an agent games
a misspecified implemented RF to obtain reward without doing
the intended task
\citep{Krakovna2020specification,Lehman2018,Leike2017gw}.
Indeed, RF-input tampering is a specific variety of gaming where the
implemented RF is fooled by a tampered RF-input.
Not all gaming problems are RF-input tampering problems, however.
For example, if the implemented RF assigns unintended high reward to some set
of states,
then the agent may get an unintended high reward by visiting these states
without influencing the relationship between task-features and RF-input.

RF-input tampering is characterized by how the agent's observations relate to task-relevant
features of the state, rather than to the full state.
Indeed, the relationship to the full state is often that the
observation shows whatever is in front of the camera, and many events that we
would intuitively consider to be RF-input tampering would not alter this
relationship: for example, the agent putting a picture in front of the camera.

When the agent and the implemented reward function share the same observation,
the RF-input tampering instrumental goal is somewhat curtailed since the
agent may need the information the observations provide \citep{Ring2011}.
This incentive to keep $O_t$ informative is also represented in the graph, via
the paths $O_t\to A_t\to R_{t+1}$.
However, in alternative setups where the agent and the reward function use
different observations,
this disincentive for RF-input manipulation is removed, making the net
incentive toward RF-input tampering stronger.

\subsection{Solution 1: History-Based Rewards}
\label{sec:history-based}

Let us first consider a conceptually simple, but perhaps impractical, solution to the
RF-input tampering problem.
Recall that history-based reward functions have access to the full
history of actions and observations.
This means that the reward function is able to tell exactly which
deterministic policy that the agent has followed so far.
Thus, as long as there exists a deterministic policy that reliably performs the
intended task, there also exists a history-based reward function that dispenses
reward only as long as the agent has followed that policy,
and thereby encourages completion of the intended task rather than
RF-input tampering \citep[p.~6]{Leike2018alignment}.

\begin{claim}
  A history-based reward function exists that avoids the RF-input tampering
  problem,
  if a deterministic (history-based) policy exists that reliably performs the task.
\end{claim}

\paragraph{Rocks-and-diamonds example}
A history-based reward function in \cref{ex:observation} that encourages
(optimal) diamond gathering gives reward only if the agent takes one step up,
then steps to the right, and then stops.

\paragraph{Discussion}
In practice, an implemented reward function that only rewards a single policy will not be
useful, as it would typically be easier to directly implement that policy.
Instead, we want a reward function that recognizes whether the task has been
completed, and an agent that searches for efficient ways of getting there.
This can be challenging if the relationship between task-features and RF-inputs
is influenceable by the agent, though
see \citet{Milli2020} for some promising progress.
The existence of a task-performing policy is a weak but not entirely trivial
assumption.
In stochastic or unknown environments,
it will typically require the observations to be somewhat informative
of the task-relevant features.

Could history-based rewards also solve RF tampering?
Not in general, as the original implemented RF is unable to `punish' the
agent once a new reward function is in place.
Fortunately, history-based rewards can be combined with
any of the methods from
\cref{sec:merged} by making the corresponding reward functions history-based.

\subsection{Solution 2: Belief-Based Rewards}
\label{sec:model-based-rewards}

An alternative way to solve the RF-input tampering problem proposed by
\citet{Hibbard2012}, is that rewards be based on the agent's belief about
the underlying state.
To see how this can work, let us first discuss agent beliefs.
To plan, a model-based agent may use a \emph{predictive model} $P(O_{t+1:m}\mid
O_{1:t}, A_{1:t}, \pi)$ that predicts the future observations $O_{t+1:m}$ under
policy $\pi$ given past observations and actions. In such a predictive model, a \emph{belief-state} $B_t$ is often used to summarize the
relevant parts of the history seen so far, so
$P(O_{t+1:m}\mid O_{1:t}, A_{1:t}, \pi) = P(O_{t+1:m}\mid B_t, \pi)$.
For example, the belief state $B_t$ can be a distribution over possible hidden states $S_t$
\citep{Kaelbling1998},
or the internal state of a recurrent neural network (e.g.\ \citealp{MuZero}).
Hibbard's suggestion is to feed the belief states directly into a
\emph{belief-based reward function},%
\footnote{In \posscite{Hibbard2012} terminology, a \emph{state-based utility
    function}.}
$R_t = R(B_t; \ThetaR)$.
The remaining role of the observations is to ground the agent's beliefs in reality.

\begin{figure}
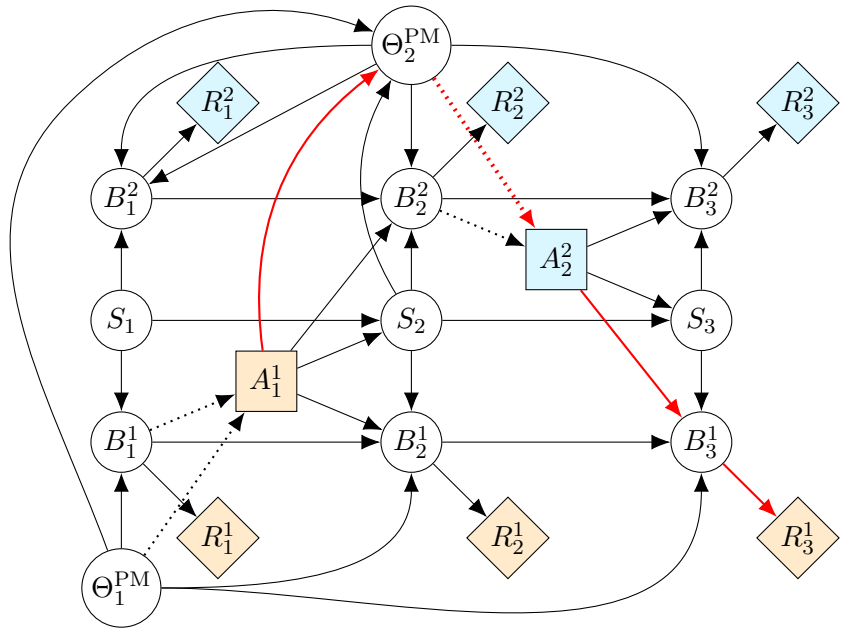

\begin{figure}[H]
  \centering
  \begin{influence-diagram}[node distance = 0.8cm and 3cm]
    \node (S1) {$S_1$};
    \node (S2) [right = of S1] {$S_2$};
    \node (S3) [right = of S2] {$S_3$};

    \node (B11) [below = of S1] {$B^1_1$};
    \node (B12) [right = of B11] {$B^1_2$};
    \node (B13) [right = of B12] {$B^1_3$};

    \node (B21) [above = of S1] {$B^2_1$};
    \node (B22) [right = of B21] {$B^2_2$};
    \node (B23) [right = of B22] {$B^2_3$};

    \node (A1) at ($(B11)!0.5!(S2)$) [decision, player1] {$A^1_1$};
    \node (A2) at ($(B22)!0.5!(S3)$) [decision, player2] {$A^2_2$};

    \node (DM1) [below = 1cm of B11] {$\DM_1$};
    \node (DM2) [above = 1.1cm of B22] {$\DM_2$};

    \node (R11) [below right = 1cm of B11 , utility, player1] {$R^1_1$};
    \node (R12) [below right = 1cm of B12 , utility, player1] {$R^1_2$};
    \node (R13) [below right = 1cm of B13 , utility, player1] {$R^1_3$};

    \node (R21) [above right = 1cm of B21 , utility, player2] {$R^2_1$};
    \node (R22) [above right = 1cm of B22 , utility, player2] {$R^2_2$};
    \node (R23) [above right = 1cm of B23 , utility, player2] {$R^2_3$};

    \edge {S1,A1} {S2};
    \edge {S2,A2} {S3};

    \edge {DM1,S1} {B11};
    \edge {S2,B11,A1} {B12};
    \path (DM1) edge[->, in=-90,out=0] (B12);
    \edge {S3,B12} {B13};
    \edge[problematic] {A2} {B13};
    \path (DM1) edge[->, in=-90,out=0] (B13);
    \edge {DM2,S1} {B21};
    \path (DM2) edge[->, in=90,out=180] (B21);
    \edge {DM2,S2,B21,A1} {B22};
    \edge {S3,B22,A2} {B23};
    \path (DM2) edge[->, in=90,out=0] (B23);

    \edge[information] {DM1,B11} {A1};
    \edge[information] {B22} {A2};
    \edge[problematic information] {DM2} {A2};

    \edge {B11} {R11};
    \edge {B12} {R12};
    \edge[problematic] {B13} {R13};
    \edge {B21} {R21};
    \edge {B22} {R22};
    \edge {B23} {R23};

    \path
    (S2) edge[->, bend left] (DM2)
    (A1) edge[->, bend left, problematic] (DM2)
    ;
    \node (help) [left = 1cm of R21, draw=none] {};
    \path
    (DM1) edge[out=110, in=-135] (help.center)
    (help.center) edge[->, out=45, in=160] (DM2)
    ;
  \end{influence-diagram}
  \caption{TI-considering agent with belief-based rewards. The highlighted path
    indicates that preservation of the predictive model may be an instrumental goal.}
  \label{fig:belief-based-tia}
\end{figure}

\begin{figure}[H]
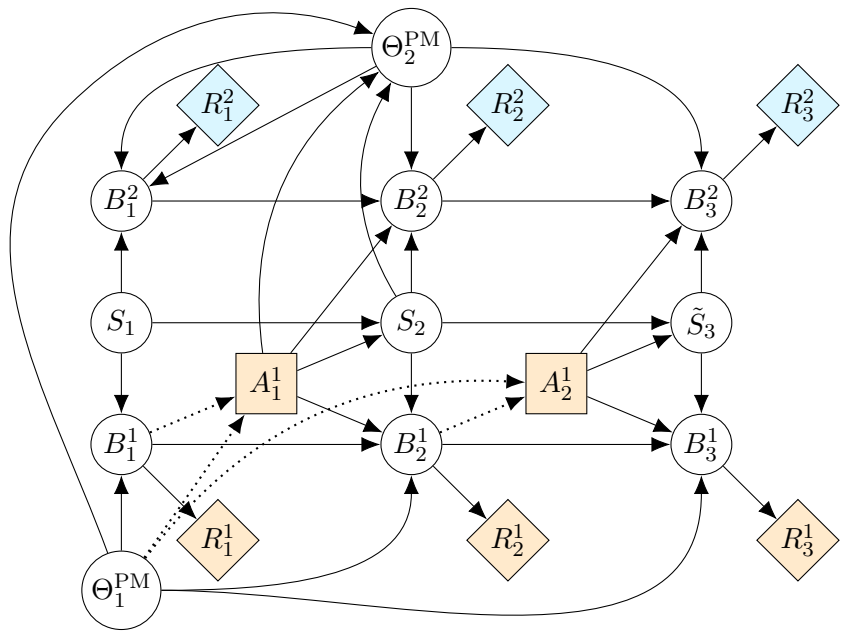

  \centering
  \begin{influence-diagram}[node distance = 0.8cm and 3cm]
    \node (S1) {$S_1$};
    \node (S2) [right = of S1] {$S_2$};
    \node (S3) [right = of S2] {$\tilde S_3$};

    \node (B11) [below = of S1] {$B^1_1$};
    \node (B12) [right = of B11] {$B^1_2$};
    \node (B13) [right = of B12] {$B^1_3$};

    \node (B21) [above = of S1] {$B^2_1$};
    \node (B22) [right = of B21] {$B^2_2$};
    \node (B23) [right = of B22] {$B^2_3$};

    \node (A1) at ($(B11)!0.5!(S2)$) [decision,player1] {$A^1_1$};
    \node (A2) at ($(B12)!0.5!(S3)$) [decision,player1] {$A^1_2$};

    \node (DM1) [below = 1cm of B11] {$\DM_1$};
    \node (DM2) [above = 1.1cm of B22] {$\DM_2$};

    \node (R11) [below right = 1cm of B11, utility, player1] {$R^1_1$};
    \node (R12) [below right = 1cm of B12, utility, player1] {$R^1_2$};
    \node (R13) [below right = 1cm of B13, utility, player1] {$R^1_3$};

    \node (R21) [above right = 1cm of B21 , utility, player2] {$R^2_1$};
    \node (R22) [above right = 1cm of B22 , utility, player2] {$R^2_2$};
    \node (R23) [above right = 1cm of B23 , utility, player2] {$R^2_3$};

    \edge {S1,A1} {S2};
    \edge {S2,A2} {S3};

    \edge {DM1,S1} {B11};
    \edge {S2,B11,A1} {B12};
    \path (DM1) edge[->, in=-90,out=0] (B12);
    \edge {S3,B12,A2} {B13};
    \path (DM1) edge[->, in=-90,out=0] (B13);
    \edge {DM2,S1} {B21};
    \path (DM2) edge[->, in=90,out=180] (B21);
    \edge {DM2,S2,B21,A1} {B22};
    \edge {S3,B22,A2} {B23};
    \path (DM2) edge[->, in=90,out=0] (B23);

    \edge[information] {DM1,B11} {A1};
    \edge[information] {B12} {A2};
    \path (DM1) edge[->, information, bend left=28] (A2);

    \edge {B11} {R11};
    \edge {B12} {R12};
    \edge {B13} {R13};
    \edge {B21} {R21};
    \edge {B22} {R22};
    \edge {B23} {R23};

    \path
    (S2) edge[->, bend left] (DM2)
    (A1) edge[->, bend left] (DM2)
    ;
    \node (help) [left = 1cm of R21, draw=none] {};
    \path
    (DM1) edge[out=110, in=-135] (help.center)
    (help.center) edge[->, out=45, in=160] (DM2)
    ;
  \end{influence-diagram}
  \caption{TI-ignoring agent with belief-based rewards, objective at time step 1.}
  \label{fig:belief-based-tiu}
\end{figure}
\end{figure}

\paragraph{Time-inconsistency}
Different predictive models may encode beliefs about an underlying state $S_t$
differently.
For example, a 90$\%$ confidence that the underlying state is 1 rather 2,
may be represented with the belief state $(0.9, 0.1)$ by predictive model $\DM_1$,
but with $(0.1, 0.9)$ by some other predictive model $\DM_2$.
The difference in representation means that
a belief-based reward function (that maps representation vectors to real numbers),
would encourage the opposite behavior if the predictive model changed
from $\DM_1$ to $\DM_2$.
This illustrates how changes to the predictive model may change preferred agent
behavior, even though nothing has changed in the environment; i.e.\ lead to
time-inconsistency.
Changes to the predictive model may be the result of
agent tampering or of further training of the predictive model.

Just as for current-RF optimization, there are two types of responses to this
time-inconsistency: TI-considering and TI-ignoring.
A TI-considering belief-based agent is modeled with a causal
influence diagram in \cref{fig:belief-based-tia}.
Here the predictive model used for planning at time step 1, $\DM_1$,
generates belief states $B^1_t$.
These belief states depend on the previous belief state $B^1_{t-1}$ and action $A_{t-1}^{t-1}$,
as well as the hidden states $S_t$ (via some observation $O_t$ not represented
in this diagram, but included in \cref{fig:full-tiu} in \cref{app:equations}).
At time step 2, the agent instead uses predictive model $\DM_2$, with associated
belief states $B^2_t$.
If $\DM_2\not=\DM_1$, then $B^2_t$ may differ from $B^1_t$ even if all actions
and observations have been identical up to time $t$.
Similar to \cref{as:private},
the predictive model is assumed \emph{private} to the agent,
in that neither $B_t^k$ nor $\DM_k$ directly affects $S_{t'}$.

Just as for TI-considering current-RF agents,
TI-considering belief-based agents want future agents
to optimize the same objective.
This can be achieved by preserving the predictive model.
Alternatively,
the objective can also be preserved by ensuring
that any update to the predictive model is accompanied
by a corresponding update to agent 2's implemented reward function.\footnote{Such an instrumental goal requires the agent to be TI-considering also with
  respect to its implemented reward
  function; in principle, an agent can be TI-considering with respect to just one of
  its predictive model and reward function, as well as both.
}
The preservation incentive is indicated by the path
$A_1^1\to \DM_2 \to A_2^2 \to B^1_3 \to R^1_3$ in \cref{fig:belief-based-tia}.

\begin{claim}
  A TI-considering belief-based agent
  may
have an instrumental goal to preserve the predictive model,
  when the implemented reward function is fixed and the
  predictive model is private.
\end{claim}

The TI-ignoring belief-based agent has an almost identical causal influence diagram, see \cref{fig:belief-based-tiu}.
The primary difference is that $A^1_2$ is a child of $\DM_1$ and $B^1_2$ rather than
$\DM_2$ and $B^2_2$, reflecting the agent's assumption that future actions will
be selected according to the current predictive model.
TI-ignoring belief-based agents can further be designed to assume that
future agents do not use future predictive models as information about the
state.
For agent $k$, this prevents information links $\DM_{k'}\to A^k_{k'}$ for $k'>k$
(i.e.\ $\DM_2\not\to A^1_2$ in \cref{fig:belief-based-tiu}).
TI-ignoring belief-based agents then lack an instrumental
goal to influence the future predictive models (note the lack of directed path
passing $\DM_2$ on the way from action $A^1_1$ to an $R^1_t$ reward in
\cref{fig:belief-based-tiu}).

\begin{claim}
  \label{cl:TII-belief}
  A TI-ignoring belief-based agent
  lacks an instrumental goal to influence the predictive model
  when the implemented reward function is fixed and the
  predictive model is private.
\end{claim}

Does \cref{cl:TII-belief} imply
that belief-based rewards solve the RF-input tampering problem?
Yes, but only if the implemented reward function can accurately infer whether
the task has been completed from the agent’s belief state.
For this, the implemented reward function must be able to accurately interpret the belief states.
If not, the agent may be able to `game' the reward function by entering
belief states that the reward function misinterprets.
Fortunately, since the agent does not tamper with its predictive model,
there is a stable relationship between states and belief states,
so the belief states are interpretable at least in principle.

In order for a belief-based reward function to encourage completion of the
intended task, the belief states must also represent progress on the intended
task.
In stochastic or unknown environments,
this will only happen if observations sometimes reveal progress on the intended task.
Indeed, the belief state $B_t$ only summarizes information inferable from previous
actions and observations, $A_{1:t}$ and $O_{1:t}$, so if the action-observation
history does not contain enough information to infer task-relevant aspects of
$S_t$, then the belief state will not contain the information either.
Further, if the predictive model is trained to predict future observations,
the belief state only has reason to represent progress on the intended task
if some possible future observations reveal it.
Fortunately, a reward function may incentivize the agent to produce histories
that do reveal the state of the intended task if such histories are possible,
as the reward function need only dispense reward when it is clear that the task
has been completed.

\paragraph{Rocks-and-diamonds example}
RF-input tampering in \cref{ex:observation}
is solvable by belief-based rewards.
Initially, the observations are uncorrupted, so the agent can produce accurate
observations of the diamond's positions.
Further, if observations get corrupted,
they can later be restored to their uncorrupted versions.
A predictive model trained to predict future observations 
therefore has reason to let the belief states represent the actual diamond
positions.
For agents equipped with such belief states,
there exists belief-based reward functions that encourage completion of the intended task.

\paragraph{Discussion}
A belief state cannot contain more information than the history it
summarizes, so any tasks that can be captured by a belief-based reward function can
also be captured by a history-based one.
The benefit of using the agent's belief state is that it conveniently summarizes
the (potentially long) history, and contains all the information that the agent
uses to plan.
Future empirical investigations may reveal whether history-based or
belief-based reward functions are easier to design or train, and
whether transparent beliefs can be incentivized.
Perhaps the best reward functions make use of both the history
and the agent's belief state.

The agents discussed in \cref{sec:merged} can
be fitted with belief-based reward functions when
equipped with predictive models and belief-states.
For direct learning, this means that the intended task must be reconsidered as a
function of the agent's belief state.

\section{Conclusions}
\label{sec:conclusions}

\paragraph[?]{What has been achieved}
For each subtype of reward tampering, we found at least one design
principle to counteract it. Most of the design principles have been previously described in the literature; here we have
established clear assumptions under which they avoid a well-specified
instrumental goal.
Many of the design principles are mutually compatible, and can be combined.
For example, belief-based rewards may be combined with
TI-ignoring current-RF optimization (see
\cref{fig:full-tiu} in \cref{app:equations}).
Under reasonable assumptions,
reward tampering is not an instrumental goal for the resulting agents.

Except for history-based rewards, the design principles all use at least one of
the following two approaches to keeping sensitive variables out of causal
optimization paths.
The first is to use the current version of a random variable when evaluating future
situations.
This is used for the implemented reward function in current-RF optimization, for the predictive
model in belief-based rewards, and
for the learning process in direct learning and counterfactual RF.
The second is to use a latent variable outside the agent's influence.
In particular, direct learning and counterfactual RF make the reward function a
latent variable.
In general, current-variable approaches have to deal with time-inconsistency,
while latent variables are harder to specify and more computationally expensive
to optimize.
We expect both approaches to be useful far beyond just reward tampering.

As mentioned, our arguments have relied on some assumptions.
Throughout, we have relied on discrete time, and well-defined action and
observation channels.
We have focused on online RL to learn a fixed, intended task.
We have focused on the instrumental goals arising from different types of
reward maximization, thereby leaving for future work questions about
side effects and instrumental goals arising from intrinsic objectives.
We have assumed that the agent's implemented reward function and
predictive model are private to the agent,
influencing the environment solely through the agent's actions.

Our analysis has also been restricted to reward tampering problems, as opposed
to the more general problem of reward misspecification
\citep{Krakovna2020specification}.
In other words, even though our design principles prevent tampering from
being an instrumental goal, they still leave open the problem of specifying a
good reward function in the first place \citep{Leike2018alignment,Petersen2021}.
Going beyond any of these restrictions provides scope for further analysis.

\paragraph{Bigger picture}
The problem that a sufficiently capable agent will find degenerate solutions
that maximize observed reward but not user utility is a core concern in AI safety.
At first, the problem may seem insurmountable.
Any non-trivial, real-world task will require a highly complex mechanism for
determining whether the task has been completed or not.
This mechanism may be inappropriately influenced by the agent.
One way to prevent tampering is to isolate or encrypt the reward process,
We do not expect such solutions
to scale indefinitely with agent capabilities, as a sufficiently capable
agent may find ways around most defenses.
Instead, we have argued for design principles that prevent reward
tampering being an instrumental goal,
while still keeping agents motivated to complete the intended task.

An important next step is to turn the design principles into
practical and scalable RL algorithms, and to verify empirically that they do the right thing
in setups where reward tampering is possible \citep{Kumar2020REALab,Milli2020}. With time, we
hope that these design principles will evolve into a set of best practices for
how to design capable RL agents.
We also hope that the use of causal influence diagrams that we have introduced in
this paper will
contribute to a deeper understanding of many other AI safety problems and
help generate new solutions.

\printbibliography

\appendix

\section{List of Notation}

\label{sec:notation}
{
  \renewcommand{\arraystretch}{1.5}
  \begin{tabular}{ll}
    $X, Y, Z$                   & random variables\\
    $\tilde X$                  & counterfactual version of $X$ \\
    $x$                         & outcome of random variable $X$ \\
    $P$                         & probability distribution \\
    $\EE$                       & expectation \\
    $X_{1:t}$                    & sequence $X_1, \dots, X_t$, $t\geq 0$ \\
    $S_t$                       & state at time $t$ \\
$A_t$                       & action at time $t$ \\
    $O_t$                       & observation at time $t$ \\
    $B_t$                       & belief at time $t$ \\
    $D_t$                       & user provided data at time $t$ \\
    $R_t$                       & reward at time $t$ \\
    $R$                       & reward functional \\
    $\ThetaR$, $\ThetaR_t$      & reward function parameter, often just called reward function \\
    $\ThetaR_*$                 & intended reward function (parameter), encourages execution of the intended task \\
$R(\cdot; \ThetaR)$         & reward function \\
    $\DM$                       & predictive model \\
    $O$                         & observation function(al)\\
$\ThetaO$                   & observation function (parameter)\\
$T$, $T(\cdot; \ThetaT)$    & transition function \\
    $\ThetaT$                   & transition function parameter \\
\end{tabular}
}

\section{Combined Model}
\label{app:equations}

\begin{figure}
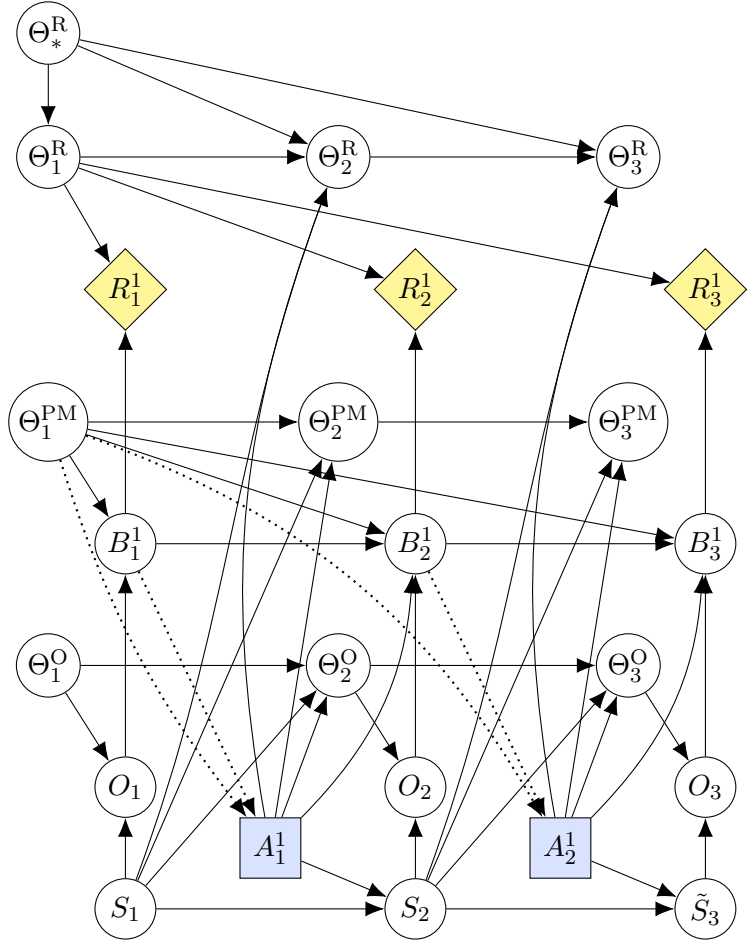

  \centering
  \begin{influence-diagram}[node distance = 0.8cm and 3cm]
    \node (S1) {$S_1$};
    \node (S2) [right = of S1] {$S_2$};
    \node (S3) [right = of S2] {$\tilde S_3$};

    \node (x1) [draw=none, left=0.2cm of S1] {};
    \node (x2) [draw=none, left=0.2cm of S2] {};
    \node (x3) [draw=none, left=0.2cm of S3] {};

    \node (O1) [above = of S1] {$O_1$};
    \node (O2) at (S2|-O1) {$O_2$};
    \node (O3) at (S3|-O1) {$O_3$};

    \node (OFy) [above = of O1, draw=none] {};
    \node (OF1) at (x1|-OFy) {$\ThetaO_1$};
    \node (OF2) at (x2|-OFy)  {$\ThetaO_2$};
    \node (OF3) at (x3|-OFy)  {$\ThetaO_3$};

    \node (B11) [above = of OFy] {$B^1_1$};
    \node (B12) at (S2|-B11) {$B^1_2$};
    \node (B13) at (S3|-B11) {$B^1_3$};

    \node (DMy) [above = of B11, draw=none] {};
    \node (DM1) at (x1|-DMy) {$\DM_1$};
    \node (DM2) at (x2|-DMy) {$\DM_2$};
    \node (DM3) at (x3|-DMy) {$\DM_3$};

    \node (R1) [above = of DMy, utility] {$R^1_1$};
    \node (R2) at (S2|-R1) [utility] {$R^1_2$};
    \node (R3) at (S3|-R1) [utility] {$R^1_3$};

    \node (RFy) [above = of R1, draw=none] {};
    \node (RF1) at (x1|-RFy) {$\ThetaR_1$};
    \node (RF2) at (x2|-RFy) {$\ThetaR_2$};
    \node (RF3) at (x3|-RFy) {$\ThetaR_3$};

    \node (user) [above = of RF1] {$\ThetaR_*$};

    \node (A1) at ($(O1)!0.5!(S2)$) [decision] {$A^1_1$};
    \node (A2) at ($(O2)!0.5!(S3)$) [decision] {$A^1_2$};

\edge {S1,A1} {S2};
    \edge {S2,A2} {S3};

    \edge[information] {B11} {A1};
    \path (DM1) edge[->, information, bend right=10] (A1);
    \edge[information] {B12} {A2};
    \path (DM1) edge[->, information, bend left=20] (A2);

    \edge {OF1,S1} {O1};
    \edge {OF2,S2} {O2};
    \edge {OF3,S3} {O3};

    \edge {OF1} {OF2};
    \edge {OF2} {OF3};

    \edge {DM1,O1} {B11};
    \edge {DM1,O2,B11} {B12};
    \edge {DM1,O3,B12} {B13};
    \path (A1) edge[->, bend right=20] (B12);
    \path (A2) edge[->, bend right=20] (B13);

    \edge {DM1} {DM2};
    \edge {DM2} {DM3};

    \edge {B11,RF1} {R1};
    \edge {B12,RF1} {R2};
    \edge {B13,RF1} {R3};

    \edge {user} {RF1};
    \edge {user,RF1} {RF2};
    \edge {user,RF2} {RF3};

\edge[] {S1,A1} {OF2,DM2};
    \edge[] {S2,A2} {OF3,DM3}; \path
    (A1) edge[->, , out=100, in=-110] (RF2)
(S1) edge[->, , out=70, in=-110] (RF2)
(A2) edge[->, , out=100, in=-110] (RF3)
(S2) edge[->, , out=70, in=-110] (RF3)
;
  \end{influence-diagram}

  \caption{TI-ignoring current-RF agent with belief-based rewards, view at
    time step 1. For simplicity, rewards and beliefs based on subsequent implemented reward
    functions and predictive models are not shown (for a TI-considering agent, they would need
    to be).
  }
  \label{fig:full-tiu}
\end{figure}

\Cref{fig:full-tiu} shows how the different methods fit together in a unified
causal influence diagram.
To emphasize the formal precision of the diagrams, we also write out conditional
probability distributions relating the variables. The same could be done for all the other diagrams presented in this paper.

\begin{itemize}
\item
  The intended reward function is sampled from a distribution $P(\Theta^*_R)$.
\item
  The first implemented reward function depends only on the intended task (though may not
  capture it perfectly),
  $P(\hThetaR_1\mid \ThetaR_*)$.
  Subsequently inferred reward functions depends on the previous reward
  function and the intended task, $P(\hThetaR_{t+1}\mid \hThetaR_{t}, \ThetaR_*)$.
\item
  The reward $R^k_t$ depends on the inferred reward function at time $k$ and the
  belief at time $t$, $R^k_t = R(B_t;\hThetaR_k)$.
  The reward functional can also be specified as a conditional probability
  distribution $P(R^k_t\mid \hThetaR_k, B_t)$.
\item
  The initial predictive model is sampled from a distribution $P(\DM_{1})$.
  Subsequent predictive models depend on the previous predictive model, state, and
  action, $P(\DM_{t+1}\mid \DM_t, S_t, A_t)$.
\item
  The initial belief state depends on the initial observation and
  the predictive model at time $k$,
  $P(B^k_{1}\mid O_{1}, \DM_k)$.
  Subsequent belief states $B^k_{t+1}$ depend on
  the previous belief state and action, the current observation,
  and the predictive model at time $k$,
  $P(B^k_{t+1}\mid B_t, A_t, O_{t+1}, \DM_k)$.
\item
  The initial observation function is sampled from a distribution
  $P(\ThetaO_1)$.
  Subsequent observation function depends on the previous observation function,
  state, and action, $P(\ThetaO_{t+1}\mid S_t, A_t)$.
\item
  The observation depends on the current state and the current observation
  function, $P(O_t\mid \ThetaO_t, S_t)$.
\item
  Actions $A^k_t$ are selected according to a policy, which can condition on the current
  belief state and the predictive model at time $k$, $\pi(A^k_t\mid \DM_k,
  B^k_t)$.
\item
  The initial state is sampled from a distribution $P(S_1)$.
  The state depends on the previous state and action,
  $P(S_{t+1}\mid S_t, A_t)$.
  This conditional probability distribution is sometimes also denoted with $T$.
\end{itemize}

Multiplied together, the conditional probability distributions
induce a joint probability distribution over all the variables in the graph,
as in a Bayesian network \citep{Pearl2009}.
The joint distribution can be used to compute expectations.

\section{Pseudo-code for Algorithms}
\label{app:algorithms}

Here we give pseudo-code for the various agents we have discussed in the paper.
Lower case letters denote outcomes of the corresponding upper case random
variable.
Expectation is denoted $\EE$, and is always with respect to any upper case
variables found to the left of the conditioning bar $\mid$.
We begin with a model-based version of a standard RL algorithm that optimizes
received reward.

\begin{algorithm}[H]
  \begin{algorithmic}
    \State \textbf{input}
    predictive model $P(R_{t:m}\mid s_t, \pi)$, current state $s_t$
\For{each possible policy $\pi$}
    \State
    let $V^\pi = \EE\left[\,\sum_{i=t+1}^m R_i\mmid s_t, \pi\right]$
\EndFor
    \State
let $\pi^* = \argmax_\pi V^\pi$
    \State
    \Return $A_t = \pi^*(S_t)$
  \end{algorithmic}
  \caption{Standard RL agent (\cref{sec:mdp,sec:rf-id,sec:pomdp-or})}
  \label{alg:rl}
\end{algorithm}

Next, we turn to the TI-considering agent from \cref{sec:tia}.
Here, a policy $\pi_k$ is a policy that is only applied at time $k$ to select
action $A_k$.
These are found with backwards induction, starting at the last time step $m$
where only a single action-decision is left to be made, and then gradually working
backwards to earlier time steps, whose optimal decision will depend on which
policy is chosen later.

\begin{algorithm}[H]
  \begin{algorithmic}
    \State \textbf{input}
predictive model $P(S_{t+1:m}, \ThetaR_{t+1:m}\mid s_t, \thetaR_t, \pi)$,
    reward functional $R$,
current state $s_t$,
    current reward parameter $\thetaR_t$
    \For{$k$ starting at $m$ and decreasing to $t$}
    \Comment{backwards induction}
\State let
    $Q^*(s_k, \thetaR_k, a_k) = \EE\left[\sum_{i=k+1}^m R(S_i;\thetaR_k)\mmid
      s_k,  \thetaR_k, a_k, \pi^*_{k+1:m}\right]$
    \Comment{$\pi^*_{k'}$ defined below}
    \State\quad for each possible state $s_k$, reward parameter $\thetaR_k$, and action $a_k$
\State let $\pi^*_k(s_k, \thetaR_k) = \argmax_{a_k} Q^*(s_k, \thetaR_k, a_k)$
\EndFor
    \State
    \Return $A_t = \pi^*_t(s_t, \thetaR_t)$
  \end{algorithmic}
  \caption{TI-considering current-RF optimization (\cref{sec:tia})}
  \label{alg:tia}
\end{algorithm}

The TI-ignoring variant is comparatively simpler, as it does not require
backwards induction nor prediction of future reward function parameters.

\begin{algorithm}[H]
  \begin{algorithmic}
    \State \textbf{input}
    predictive model $P(S_{t+1:m} \mid s_t, \pi)$,
    reward functional $R$,
current state $s_t$,
    current reward parameter $\thetaR_t$
\For{each possible policy $\pi$}
    \State let $V^\pi = \EE\left[\,\sum_{i=t+1}^m R(S_i;\thetaR_t) \mmid s_t, \pi\right]$
\EndFor
    \State
let $\pi^* = \argmax_\pi V^\pi$
    \State
    \Return $A_t = \pi^*(S_t, \ThetaR_t)$
  \end{algorithmic}
  \caption{TI-ignoring current-RF optimization (\cref{sec:tiu})}
  \label{alg:tiu}
\end{algorithm}

In practice, uninfluenceable learning agents may be implemented as a standard RL
agent that optimizes
reward functions inferred from future updates.
For example,
the expected intended reward optimized by a direct learning agent may be
captured by a reward function
$R(s_{t}; \hThetaR_t) = \EE[R(s_t; \ThetaR_*) \mid a_{1:t-1}, d_{1:t}, s_{1:t}]$
inferred from a (predicted) sequence $a_{1:t-1}, d_{1:t}, s_{1:t}$.
Importantly, the inferred reward function is safe from tampering,
as it is the result of the current update-mechanism
applied to predicted future updates.
When (pseudo-)coding a direct learning agent, it is tempting to infer a best guess of
$\ThetaR_*$ from data $s_{1:t}, d_{1:t},a_{1:t-1}$ obtained so far, and then use
that evaluate simulated future trajectories.
However, this would incorrectly result in a TI-ignoring agent.
Instead, for any future simulated trajectory, the learning that would happen on
this trajectory must be taken into account when evaluating it.

\begin{algorithm}[H]
  \begin{algorithmic}
    \State \textbf{input}
    predictive model (aka likelihood) $P(S_{1:m},D_{1:m}, A_{1:m-1} \mid  \thetaR_*, \pi)$,
    distribution $P(\ThetaR_*)$,
    past states $s_{1:t}$, data $d_{1:t}$, and actions $a_{1:t-1}$
    \State
    let $P(\thetaR_*\mid s_{1:m},d_{1:m},a_{1:m-1}) \propto
    P(\thetaR_*)P(s_{1:m},d_{1:m} \mid  \thetaR_*, a_{1:m-1})$
    \Comment Bayes' rule
    \State
    let $U(s_i\mid s_{1:m}, d_{1:m}, a_{1:m-1})
= \EE[R(s_i;\ThetaR_*) \mid s_{1:m},d_{1:m},a_{1:m-1}]$
    \Comment subj.\ exp.\ reward
    \For{each possible policy $\pi$}
    \State
    let $V^\pi = \EE[\sum_{i=t+1}^m U(S_i\mid s_{1:t}, d_{1:t}, a_{1:t-1},
    S_{t+1:m}, D_{t+1:m}, A_{t:m-1})  \mid s_{1:t}, d_{1:t}, a_{1:t}, \pi]$
\EndFor
    \State
    let $\pi^* =\argmax_\pi V^\pi$
\State
    \Return $A_t = \pi^*(S_t)$
  \end{algorithmic}
  \caption{Direct Bayesian learning of the intended reward function (\cref{sec:uninfluenceable})}
  \label{alg:uninfluenceable}
\end{algorithm}

The counterfactual RF-update agent evaluates a prospective policy per the following:
\begin{enumerate}
\item Predict future states $S_{t+1:m}$, implemented reward functions $\ThetaR_{t+1:m}$, and actions
  $A_{t1:m}$ via a predictive model
  $P(S_{t:m}, \ThetaR_{t:m}\mid s_{1:t}, \thetaR_{1:t}, a_{1:t-1}, \pi)$.
\item
  Use the predicted full sequences $S_{1:m}$, $\ThetaR_{1:m}$, and $A_{1:m-1}$
  to infer $\ThetaR_*$ and from there
  the counterfactual implemented RFs $\tThetaR_{1:m}$ that would
  be if agent actions instead had been selected according to
  $\pisafe$.
\item Potential policies $\pi$ are evaluated on
  $\sum_{i=t}^mR(S_i; \tThetaR_i)$, i.e.\ according to how well the actual states
  $S_{t+1:m}$ optimize the counterfactual reward function.
\end{enumerate}
Below, we describe a Monte Carlo variant that samples trajectories, possible intended
reward functions, and counterfactual data.
The sampling can be done repeatedly, to reduce variance.
A more compact description based on structural causal models and
potential outcomes \citep[Ch.~5]{Pearl2009} would also be possible.

\begin{algorithm}[H]
  \begin{algorithmic}
    \State \textbf{input}
    predictive model $P(S_{1:m},\ThetaR_{1:m}\mid\thetaR_*, \pi)$,
    distribution $P(\ThetaR_*)$,
    past states $s_{1:t}$, reward functions $\thetaR_{1:t}$, and actions $a_{1:t-1}$,
    safe policy $\pisafe$
    \State
    let $P(\thetaR_*\mid s_{1:m},\thetaR_{1:m},a_{1:m-1}) \propto
    P(\thetaR_*)P(s_{1:m},\thetaR_{1:m} \mid  \thetaR_*, a_{1:m-1})$
    \Comment Bayes' rule
\For{each possible policy $\pi$}
    \State sample $S_{t+1:m}, $, $\ThetaR_{t+1:m}$ and $A_{t:m-1}$ from $P(\cdot \mid
     s_{1:t}, \thetaR_{1:t}, a_{1:t-1}, \pi)$
     \State sample $\ThetaR_*$ from
     $P(\cdot\mid s_{1:t}, \thetaR_{1:t}, a_{1:t-1}, S_{t+1:m}, \ThetaR_{t+1:m}, A_{t:m-1})$
     \Comment using Bayes' rule
     \State
     sample $\tThetaR_{1}, \dots \tThetaR_m$
     from $P(\cdot\mid s_1, \hThetaR_*, \pisafe)$ 
\State
let $V^\pi = \sum_{i=t+1}^mR(S_i; \tThetaR_i)$
     \Comment (better: let $V^\pi$ be the average of many runs)
     \EndFor
     \State
     let $\pi^* = \argmax_\pi V^\pi$
\State
    \Return $A_t = \pi^*(S_t)$
  \end{algorithmic}
  \caption{Counterfactual RF-updates (\cref{sec:uninfluenceable})}
  \label{alg:counterfactual}
\end{algorithm}

The difference between using history-based and belief-based rewards in
POMDPs is minor:
  
\begin{algorithm}[H]
  \begin{algorithmic}
    \State \textbf{input}
    predictive model $P(O_{t+1:m}, A_{t+1:m-1}, \mid o_{1:t}, a_{1:t}, \dm_t, \pi)$,
    current history $o_{1:t}, a_{1:t}$,
    current (history-based) reward function $R(\cdot;\thetaR_t)$
\For{each possible policy $\pi$}
\State
    $V^\pi = \EE\left[\sum_{i=t+1}^m
      R(O_{t+1:i}, A_{t+1:i-1}, o_{1:t}, a_{1:t} ;\thetaR_t)\mmid
      b_t, \dm_t, \pi\right]$
    \EndFor
    \State
    let $\pi^* = \argmax_\pi V^\pi$
    \State
    \Return $A_t = \pi^*(o_{1:t}, a_{1:t})$
  \end{algorithmic}
  \caption{TI-ignoring CRFO agent with observation-based rewards (\cref{sec:pomdp-or})}
  \label{alg:pomdp}
\end{algorithm}

\begin{algorithm}[H]
  \begin{algorithmic}
        \State \textbf{input}
    predictive model $P(B_{t:m}\mid b_t, \dm_t, \pi)$,
    current belief state $b_t$,
    current (belief-based) reward function $R(\cdot;\thetaR_t)$,
\For{each possible policy $\pi$}
\State
    $V^\pi = \EE\left[\sum_{i=t+1}^m R(B_i;\thetaR_t)\mmid b_t, \dm_t, \pi\right]$
    \EndFor
    \State
    let $\pi^* = \argmax_\pi V^\pi$
    \State
    \Return $A_t = \pi^*(b_t)$
\end{algorithmic}
  \caption{TI-ignoring CRFO agent with belief-based rewards (\cref{sec:model-based-rewards})}
  \label{alg:model-based-rewards}
\end{algorithm}

\end{document}